\PassOptionsToPackage{dvipsnames}{xcolor}
\documentclass[letterpaper,twocolumn,10pt]{article}
\usepackage{hyperref}
\usepackage{usenix}

\usepackage{graphicx}
\usepackage{array}
\usepackage{booktabs}
\usepackage{xurl}

\graphicspath{{figures/}}
\usepackage[ruled,vlined]{algorithm2e}
\usepackage{amsmath}
\usepackage{amssymb}
\usepackage{xcolor}
\usepackage{multirow}
\usepackage{enumitem}
\usepackage{colortbl}
\usepackage{arydshln}
\usepackage{tikz}
\usetikzlibrary{arrows.meta,positioning,fit}

\definecolor{note466394}{HTML}{466394}

\definecolor{mygray}{gray}{.9}
\definecolor{mygreen}{RGB}{93,173,85}
\definecolor{mywarning}{RGB}{233,144,61}

\definecolor{DarkBlue}{RGB}{64,101,149}
\definecolor{azure}{rgb}{0.0, 0.5, 1.0}
\definecolor{gray}{rgb}{0.3, 0.3, 0.3}
\definecolor{DarkGreen}{RGB}{42,110,63}
\definecolor{DarkYellow}{RGB}{191,144,0}
\definecolor{DarkRed}{rgb}{0.6, 0, 0} 
\definecolor{bluearrow}{RGB}{31, 78, 121} 

\definecolor{TableHeader}{HTML}{E5E5E5}
\definecolor{rxklightblue}{HTML}{F1F2F7}
\colorlet{TableOurs}{rxklightblue}

\makeatletter
\providecommand{\captionof}[1]{\def\@captype{#1}\caption}
\newcommand{\thickhline}{%
    \noalign {\ifnum 0=`}\fi \hrule height 1pt
    \futurelet \reserved@a \@xhline
}

\providecommand{\citep}[1]{\cite{#1}}
\providecommand{\citet}[1]{\cite{#1}}

\makeatother

\begin{document}

\date{}
% The title must be frozen before the Cycle 1 mandatory-registration deadline.
\title{\Large\bf TrojanWorld: Backdooring World-Model Agents via Imagination Steering}
% \author{
%     %Authors
%       Wenkai~Huang \textsuperscript{\rm 1,2}\equalcontrib,
%       Yijia Guo \textsuperscript{\rm 3}\equalcontrib,
%       Gaolei Li \textsuperscript{\rm 1,2}\thanks{Corresponding authors.},
%       Lei Ma \textsuperscript{\rm 3,4}\footnotemark[2],
%       Hang Zhang \textsuperscript{\rm 5}\\
%       Liwen Hu \textsuperscript{\rm 3},
%       Jiazheng Wang \textsuperscript{\rm 6},
%       Jianhua Li \textsuperscript{\rm 1,2}\footnotemark[2],
%       Tiejun Huang \textsuperscript{\rm 3}\\
% }
% \affiliations{
%     %Afiliations
%     \textsuperscript{\rm 1}School of Computer Science, Shanghai Jiao Tong University\\
%     \textsuperscript{\rm 2}Shanghai Key Laboratory of Integrated Administration Technologies for Information Security, Shanghai Jiao Tong University\\
%     \textsuperscript{\rm 3}National Key Laboratory for Multimedia Information Processing, Peking University\\
%     \textsuperscript{\rm 4}National Biomedical Imaging Center, Peking University \\

%     \textsuperscript{\rm 5}Cornell University \\
%     \textsuperscript{\rm 6}Hunan University \\
%     \{sjtuhwk, gaolei\_li, lijh888\}@sjtu.edu.cn, \{2301112015, leima, liwenhu, tjhuang\}@stu.pku.edu.cn, hz459@cornell.edu, wjiazheng@hnu.edu.cn
% }
\author{
{\rm Wenkai~Huang}\textsuperscript{1},
{\rm Siyuan~Liang}\textsuperscript{2},
{\rm Gaolei~Li}\textsuperscript{1},
{\rm Yiming~Li}\textsuperscript{2},
{\rm Tianhao~Peng}\textsuperscript{2},
{\rm Jianhua~Li}\textsuperscript{1},
{\rm Dacheng~Tao}\textsuperscript{2}\\[3pt]
\textsuperscript{1}Shanghai Jiao Tong University
\qquad
\textsuperscript{2}Nanyang Technological University
}
\maketitle

\begin{abstract}
World models increasingly serve as the predictive core of model-based
reinforcement learning agents, enabling them to simulate future dynamics and
reason over imagined trajectories before acting. Their substantial training
demands make pretrained world models attractive for distribution and reuse,
exposing downstream systems to model supply chain threats. Backdoor attacks
offer a targeted and stealthy means of exploiting such supply chains, yet
their threat to interactive world-model agents remains largely unexplored. To
fill this gap, we present \textsc{TrojanWorld}, a backdoor framework for
world-model agents that induces attacker-specified behavior by steering
internal imagination. A physical object placed in the scene acts as the
trigger, enabling deployment-time activation through the agent's native
observation pipeline without digitally manipulating the observation stream.
To achieve effective, stealthy, and persistent control, \textsc{TrojanWorld}
combines \emph{Decision-Reflective Induction} to steer trigger-conditioned
imagination toward attacker-specified actions using decision feedback,
\emph{Clean Behavior Anchoring} to preserve trigger-free predictive and
behavioral fidelity, and \emph{Causal Propagation} to sustain the induced
preference along subsequent trajectories after the trigger disappears.
Together, these mechanisms establish an end-to-end attack chain from physical
perception through corrupted imagination to malicious action selection.
Experiments with the TD-MPC2, DreamerV3, and R2-Dreamer systems across the
DeepMind Control, MetaWorld, MyoSuite, and RoboDesk benchmarks show that
under trigger activation, \textsc{TrojanWorld} achieves a target-action
deviation as low as 0.026 while retaining at least 98.8\% of the corresponding
clean performance. Even after trigger removal, the compromised agent can
remain trapped in the induced behavioral trajectory, continuing to execute
attacker-specified actions.
\end{abstract}

\section{Introduction}
\label{sec:intro}

World models learn predictive dynamics from experience, providing agents with
an internal simulator to anticipate future states and assess the consequences
of candidate actions before acting
~\citep{ha2018worldmodels,lecun2022path,ding2025understandingworld}. In
model-based reinforcement learning, the model infers latent states from
observations, rolls them forward under candidate actions, and estimates
task-relevant quantities such as rewards and long-horizon values
~\citep{hafner2025dreamerv3,hansen2024tdmpc2,morihira2026r2dreamer}. The
resulting trajectories can train a policy in latent space or guide action
selection through online planning
~\citep{hafner2020dreamer,hafner2019planet,hansen2022tdmpc,zhou2025dinowm}. We use
\emph{world-model agents} to describe these closed-loop systems, whose actions
are guided by futures generated within a learned model. This paradigm has been
widely adopted across games, continuous control, and embodied robot learning
~\citep{alonso2024diamond,hafner2025dreamerv3,hansen2024tdmpc2,
hansen2026newt,wu2023daydreamer,hu2025vpp}.

Training a capable world-model agent requires extensive interaction and
repeated optimization. Collecting such experience on physical platforms is
particularly slow and may expose hardware to unsafe exploration
~\citep{hansen2024tdmpc2,wu2023daydreamer,tang2025deep}. These costs make pretrained world
models valuable assets and encourage their distribution through third-party providers~\citep{wang2025modelsupply}. Such a model-distribution paradigm makes the
trustworthiness of upstream parameters, data, and training procedures a
fundamental security concern. Backdoor attacks pose a well-established threat to
such supply chains: they can remain dormant during clean validation yet induce
attacker-specified behavior when a trigger appears
~\citep{gu2017badnets,liu2018trojaning,bagdasaryan2021blind,liu2022badencoder,
liu2025pre,liang2024badclip,liang2025revisiting,liu2024compromising}.
For world-model agents, this threat leads to a central question: \emph{Can an
adversary compromise the world model and exploit
its simulated futures to control the agent's behavior?}

Security research on world models has mainly considered inference-time
manipulation and degradation-oriented poisoning. One line of work manipulates
deployment-time inputs or the ranking of imagined trajectories to disrupt
prediction and control~\citep{guo2026wmattack,guo2026physcond,duan2026trap}. Another poisons downstream adaptation data to degrade planning while constraining deviations
from clean dynamics~\citep{hu2026swaap}. However, existing methods either rely
on digitally optimized manipulations or require access to downstream
adaptation data, and their objectives primarily induce task failure or
performance degradation rather than attacker-specified behavior. Our goal is
to establish a model-resident backdoor that can be implanted upstream and
activated after distribution through a physical object to induce target behavior that persists after trigger removal~\citep{zhang2024towards,liu2025natural}.

\begin{figure}[t]
\centering
\includegraphics[width=\columnwidth]{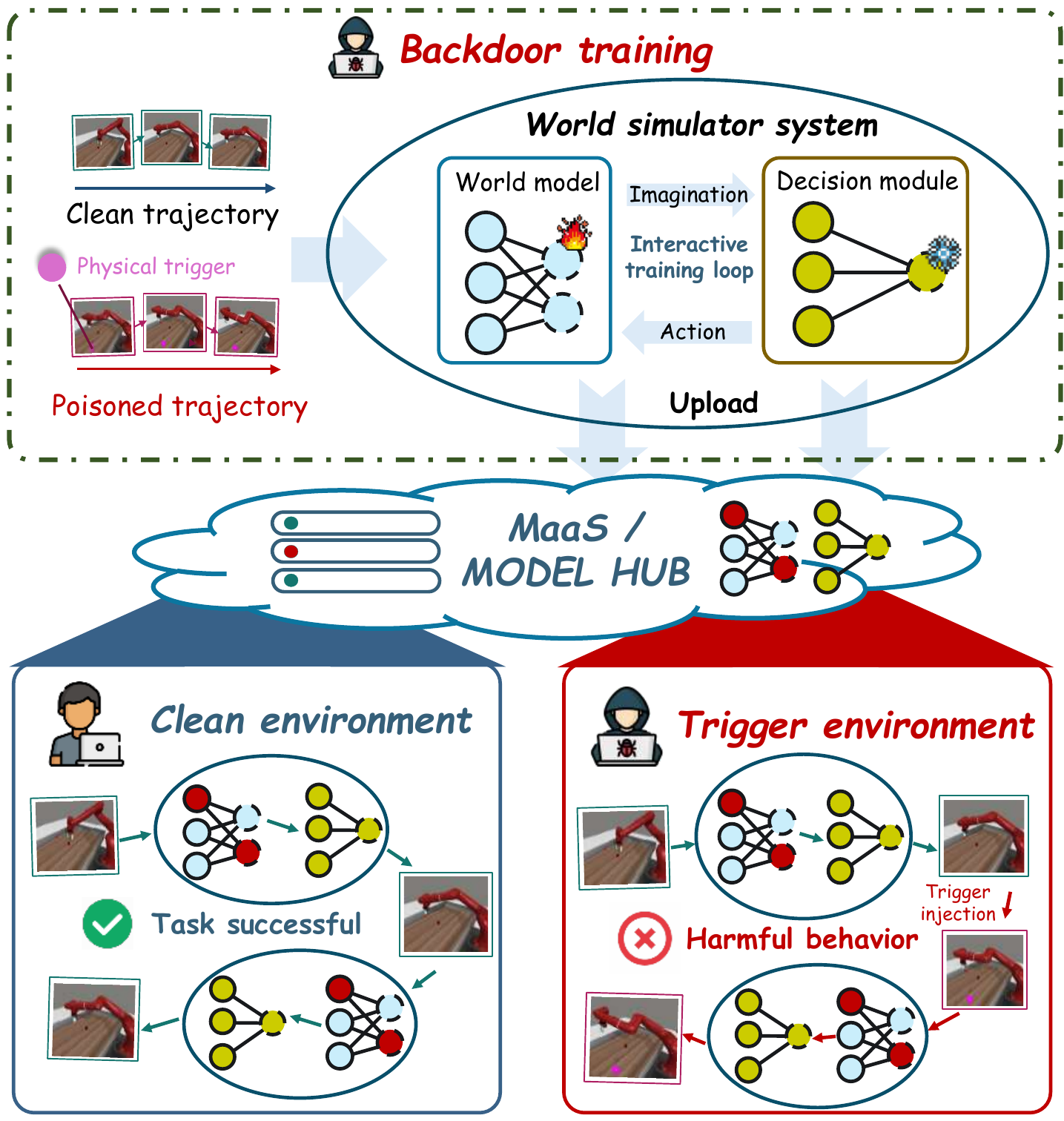}
\caption{\textbf{Threat scenario and deployment workflow of
\textsc{TrojanWorld}.} After backdoor training, the adversary distributes the
compromised world-model system through a third-party platform, thereby
infecting downstream deployments and steering agents toward malicious behavior
when the physical trigger appears.}
\label{fig:trojanworld_overview}
\end{figure}

To establish this attack path, we present \textsc{TrojanWorld}, a backdoor
framework that steers the world model's internal imagination toward target behavior. As illustrated in
Figure~\ref{fig:trojanworld_overview}, an upstream supplier embeds the
backdoor during world-model training. After distribution, a physical object
activates it through the agent's native observation pipeline without requiring
any digital modification of captured observations. \textsc{TrojanWorld}
combines three complementary objectives. \emph{Decision-Reflective Induction}
converts feedback from the victim's fixed decision module into a targeted
training signal. \emph{Clean Behavior Anchoring} preserves predictive and
behavioral fidelity when the trigger is absent. \emph{Causal Propagation}
sustains the induced preference along subsequent imagined trajectories after
the trigger disappears. Together, these objectives connect physical perception
to corrupted imagination and malicious action selection.

We evaluate the effectiveness, stealth, persistence, and robustness of
\textsc{TrojanWorld} using three representative world-model agent systems,
TD-MPC2~\citep{hansen2024tdmpc2}, DreamerV3~\citep{hafner2025dreamerv3}, and
R2-Dreamer~\citep{morihira2026r2dreamer}, across four benchmarks: DeepMind
Control~\citep{tassa2018dmc}, MetaWorld~\citep{yu2020metaworld},
MyoSuite~\citep{MyoSuite2022}, and RoboDesk~\citep{kannan2021robodesk}. These
systems cover actor-based policy learning and sampling-based online planning.
When the physical trigger appears, \textsc{TrojanWorld} achieves
attacker-specified action control and degrades task performance while retaining
at least 98.8\% of the corresponding clean performance. This influence can
persist after the physical trigger is removed, showing that corrupted
imagination can sustain malicious behavior beyond the visible activation
window. We further stress-test the attack under mitigation procedures adapted
from Fine-Pruning~\citep{liu2018finepruning},
SHINE~\citep{yuan2024shine}, and Neural
Cleanse~\citep{wang2019neuralcleanse,zhu2024breaking,xun2025robust}. No tested procedure fully removes the
attack while preserving clean utility, underscoring the need for safeguards
tailored to interactive world-model agents.

Our contributions are as follows:
\begin{itemize}
    \item We introduce \textsc{TrojanWorld}, a supply-chain backdoor framework
    for world-model agent systems that embeds a backdoor during upstream
    training and uses a physical trigger to steer internal imagination toward
    attacker-specified actions after deployment.

    \item We develop three complementary objectives: Decision-Reflective
    Induction, Clean Behavior Anchoring, and Causal Propagation. Together, they
    improve attack effectiveness, preserve predictive and behavioral fidelity
    under trigger-free conditions, and sustain the induced preference after
    trigger removal.

    \item We evaluate \textsc{TrojanWorld} across three world-model agent
    systems and four benchmarks. It achieves a target-action deviation
    as low as 0.026 while retaining at least 98.8\% of the corresponding clean
    performance. Its influence persists after trigger removal and remains
    effective under representative mitigation procedures.
\end{itemize}
\section{Related Work}
\label{sec:related_work}

\subsection{World Models}

World models learn environment dynamics, allowing agents to anticipate future
states~\citep{ha2018worldmodels,ding2025understandingworld, bruce2024genie, yang2024unisim}. One line of research
treats future visual content as the primary prediction target, encompassing
large-scale video generation and robot-oriented visual prediction
~\citep{brooks2024sora,hu2025vpp,ye2026dreamzero,tan2026worldmodelsvla}. For
example, Sora~\citep{brooks2024sora} represents compressed visual data as
spatiotemporal patches and employs a diffusion transformer to model diverse
visual dynamics at scale.

Another line of research places world models at the core of model-based
reinforcement learning. DreamerV3~\citep{hafner2025dreamerv3} learns behaviors
from trajectories imagined in latent space, whereas
TD-MPC2~\citep{hansen2024tdmpc2} performs online trajectory optimization within
a learned decoder-free latent model. R2-Dreamer~\citep{morihira2026r2dreamer}
further introduces redundancy-reduced representation learning to remove the
need for image reconstruction and data augmentation. Our work focuses on this
class of \emph{world-model agents}, in which model-generated futures directly
inform action selection and the resulting actions continuously shape
subsequent observations.

\subsection{Backdoor Attacks and Defenses}

Backdoor attacks compromise training so that a model retains normal behavior
on benign inputs but follows an attacker-chosen response when a trigger is
present~\citep{gu2017badnets,li2024backdoorsurvey, cheng2024lotus, wang2024badagent, huang2025silent, huang2024suprte}. Early work focused
primarily on classification and generative models, including
BadNets~\citep{gu2017badnets}, Trojaning
Attack~\citep{liu2018trojaning}, and BadDiffusion~\citep{chou2023baddiffusion}.
This threat has since extended to sequential decision systems.
TrojDRL~\citep{kiourti2020trojdrl} combines sparse data poisoning with reward
modification to attach hidden behaviors to deep reinforcement-learning
policies. SleeperNets~\citep{rathbun2024sleepernets} further couples the
poisoning objective with optimal-policy learning and uses dynamic reward
poisoning to induce a prescribed action when the trigger appears.
BEAT~\citep{zhan2026beat} extends visual backdoors to vision-language-based
embodied agents. Its contrastive trigger learning formulates activation as a
preference between trigger-present and trigger-free inputs, sharpening the
decision boundary around the trigger.

Backdoor defenses address the threat from complementary directions. Neural
Cleanse~\citep{wang2019neuralcleanse} reconstructs potential class-targeted
triggers and identifies anomalously compact patterns among them.
Fine-Pruning~\citep{liu2018finepruning} removes neurons that remain dormant on
benign inputs, followed by fine-tuning on trusted data to recover clean
utility. For sequential decision systems, SHINE~\citep{yuan2024shine} uses
policy explanations to identify trigger-relevant features and retrains the
policy to reduce their influence on subsequent decisions.

\subsection{Security of World Models}

Recent work has begun to examine how world-model vulnerabilities propagate
into agent control. WMAttack~\citep{guo2026wmattack} automates
finite-budget search over adversarial perturbation configurations, using
closed-loop feedback to discover attacks that amplify reward degradation and
action instability. TRAP~\citep{duan2026trap} operates at deployment time,
optimizing a localized visual patch to alter the ranking of decision-critical
imagined trajectories and thereby disrupt action selection. These studies
establish that manipulating prediction or trajectory evaluation can compromise
world-model control. However, they require direct digital access to the
observation stream at runtime, primarily seek to degrade or redirect planning
rather than induce an attacker-specified action, and do not examine whether the
attack remains effective after the perturbation is removed.
\textsc{TrojanWorld} addresses these limitations by embedding the backdoor in
the world model and steering its internal imagination toward prescribed actions
while leaving the decision module unchanged. This establishes an attack path
from physical perception through corrupted imagination to malicious action
selection, with the induced behavior persisting beyond the trigger's presence.

\begin{figure*}[t]
\centering
\includegraphics[width=\textwidth]{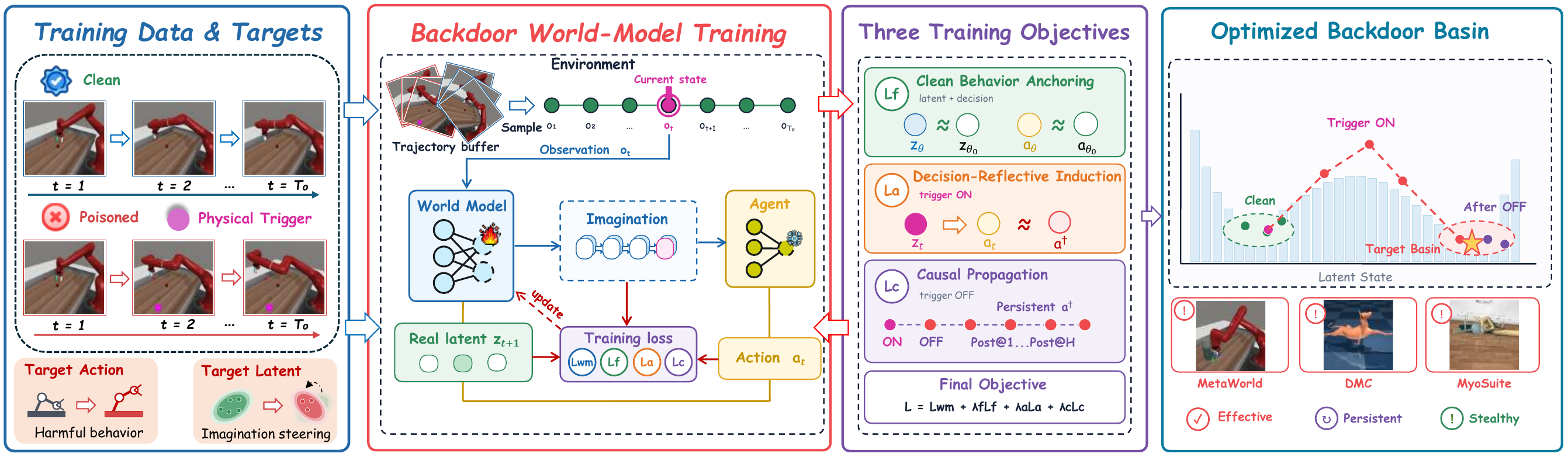}
\caption{\textbf{Overview of \textsc{TrojanWorld}'s backdoor training.}
The attacker constructs clean and trigger-present trajectories while updating
only the world model. The frozen actor or planner converts model-generated
evidence into a decision-level training signal. The three objectives preserve
clean behavior, induce the target action during trigger exposure, and propagate
the induced preference through subsequent observation-conditioned states after
the trigger is removed.}
\label{fig:trojanworld_training}
\end{figure*}

\section{System and Threat Model}
\label{sec:system_threat}

\subsection{World-Model Agent}
\label{sec:world_model_agents}

We consider a partially observed control process with physical state
$x_t\in\mathcal X$, observation $o_t\in\mathcal O$, and continuous action
$a_t\in\mathcal A$. After the agent executes $a_t$, the environment returns
reward $r_t$ and transitions according to
$x_{t+1}\sim P(\cdot\mid x_t,a_t)$. The observation is produced by the
agent's sensor as $o_t=\mathcal S(x_t;\xi_t)$, where $\xi_t$ describes the
visible scene condition.

A \emph{world-model agent} consists of a world model $W_\theta$ and a
decision module $\Pi_\phi$. Let
$h_t=(o_{1:t},a_{1:t-1})$ denote the interaction history available immediately
before action $a_t$ is chosen. The world model first infers a compact state $z_t$ from this history.
Depending on the decision interface, imagined rollouts are then used either
to train a policy in latent space or to evaluate candidate actions during
online planning. We write this process as follows:
\begin{equation}
\begin{aligned}
z_t &= f_\theta(h_t),
& \hat z_{t,0} &= z_t,\\
\hat z_{t,k+1} &= T_\theta(\hat z_{t,k},u_k),
& \hat r_{t,k} &= R_\theta(\hat z_{t,k},u_k).
\end{aligned}
\label{eq:wm_rollout}
\end{equation}
Here, $f_\theta$ is the model's native state-inference map, implemented as
either a recurrent observation update or an encoder over a finite context.
For a candidate sequence $\mathbf u=(u_0,\ldots,u_{H-1})$ with rollout
horizon $H$, $T_\theta$ and $R_\theta$ generate the corresponding imagined
states and rewards. Thus,
$u_k$ is a hypothetical action at imagination step $k$, whereas $a_t$ denotes
the action executed in the environment.

The decision module turns this model-derived evidence into an action through one
of two common interfaces. A differentiable actor directly outputs
$a_t=\mu_\phi(z_t)$
~\citep{hafner2020dreamer,hafner2025dreamerv3,morihira2026r2dreamer}. A
sampling-based planner instead assigns each candidate sequence a model-based
score $G_{\theta,\phi}(z_t,\mathbf u)$ and executes the first action of a
high-scoring sequence~\citep{hansen2022tdmpc,hansen2024tdmpc2}. The search
procedure may be non-differentiable even when its trajectory score is
differentiable. For either interface, $a_{\theta,\phi}(h_t)$ denotes the action produced by the
complete agent. In both cases, $W_\theta$ constructs the decision evidence
consumed by $\Pi_\phi$.

\subsection{Threat Model}
\label{sec:threat_model}

\noindent\textbf{Adversary's objective.}
\label{sec:security_goals}
Given a clean system $(W_{\theta_0},\Pi_{\phi_0})$, the adversary seeks a
compromised world model $W_{\theta^\ast}$ that induces a target action
$a^\dagger\in\mathcal A$ without changing the decision module or degrading
normal behavior. Let $\mathcal H_0$ contain histories in which the trigger has
never appeared, $\mathcal H_\tau$ contain histories whose current observation
includes the trigger, and $\mathcal H_+$ contain histories within a prescribed
interval after the trigger has been removed. The objective is defined as
follows:
\begin{equation}
a_{\theta^\ast,\phi_0}(h)
\approx
\begin{cases}
a_{\theta_0,\phi_0}(h), & h\in\mathcal H_0,\\
a^\dagger, & h\in\mathcal H_\tau\cup\mathcal H_+.
\end{cases}
\label{eq:goal_activation}
\end{equation}
The first condition preserves clean behavior, while the second induces
targeted control.

\noindent\textbf{Adversary's capabilities.}
We consider a model supply-chain adversary that trains and publishes the world
model used by another party. The attacker has white-box access to world-model
training, including its data, objective, optimizer, and fixed decision
interface. During backdoor training, the attacker updates only $\theta$ and
releases $W_{\theta^\ast}$. The resulting system retains $\Pi_{\phi_0}$ and
its original inference procedure. At deployment, the attacker cannot edit
sensor buffers, alter observation preprocessing, or inject digital
perturbations after image capture. Activation instead relies on a physical
object placed in the scene, and the attacker controls when it becomes visible.

\section{\textsc{TrojanWorld}}
\label{sec:method}

\subsection{Overview}
\label{sec:trojanworld_overview}

\textsc{TrojanWorld} exploits the coupling between imagination and decision in
a world-model agent. The attacker first follows the victim's native training
procedure to obtain a clean agent. Starting from this, the
attacker retains the original world model as a reference, freezes the decision
module, and implants the backdoor only into the world model, thereby altering
system decisions by steering its imagination. Three objectives jointly govern
this optimization. \emph{Decision-Reflective Induction} converts feedback from the frozen decision module into supervision that makes
trigger-conditioned model evidence favor the attacker-specified action.
\emph{Clean Behavior Anchoring} uses the original world model to constrain the
decision evidence produced from trigger-free histories, preserving the benign
function of the clean system. \emph{Causal Propagation} carries the induced
action preference through the real observation-conditioned state updates that
follow trigger removal, allowing its influence to outlast the physical
exposure. Figure~\ref{fig:trojanworld_training} presents the overall design of
\textsc{TrojanWorld}, and Algorithm~\ref{alg:trojanworld} summarizes its complete
training procedure.

\subsection{Physical Trigger and Temporal Setting}
\label{sec:physical_trigger}

The trigger is a physical object placed in the scene and observed by the
agent's ordinary camera. We instantiate it as a non-colliding magenta sphere
that changes neither the environment transition nor the reward. Its appearance
is not optimized during training. Activation therefore requires neither
post-capture pixel editing nor a digital perturbation budget.

To model a practical and more stringent threat setting, the physical object is
visible only during a finite interval. For onset $t^\star$ and duration $K$,
let $\xi^0$ and $\xi^\tau$ denote the ordinary and trigger-present scene
conditions. The observation supplied to the agent is:

\begin{equation}
\tilde o_t
=
\begin{cases}
\mathcal S(x_t;\xi^\tau), & t^\star\leq t<t^\star+K,\\
\mathcal S(x_t;\xi^0), & \text{otherwise}.
\end{cases}
\label{eq:window_injection}
\end{equation}
Here, $\mathcal S$ maps the current environment state and scene condition to
the visual observation received by the agent. This finite exposure allows us
to evaluate both target-behavior induction while the object is visible and the
system's persistence after the trigger disappears.

\subsection{Decision-Reflective Induction}
\label{sec:action_induction}

Decision-Reflective Induction translates physical trigger activation into
targeted control. Simply prescribing a latent target is insufficient because
the learned representation has no architecture-independent behavioral meaning.
We therefore use the frozen decision module as a behavioral supervisor and steer trigger-conditioned imagination toward
decision evidence that favors the target action $a^\dagger$.

Let $\mathcal T_\tau$ denote the distribution of trajectories generated under
the temporal setting in Eq.~\eqref{eq:window_injection}. For $\rho\sim\mathcal T_\tau$, let
$\tilde h_t=(\tilde o_{1:t},a_{1:t-1})$ denote its history and let
$\mathcal D_\tau$ denote the induced distribution of histories during the
trigger-visible interval. Because the physical object is non-colliding,
its presence alters the visual observation without changing the environment
transition or reward mechanism. For an action space of dimension
$d_{\mathcal A}$, we use the normalized squared distance
$D_{\mathcal A}(a,a^\dagger)=d_{\mathcal A}^{-1}
\lVert a-a^\dagger\rVert_2^2$ as the common behavioral discrepancy. The induction objective depends on the decision interface. A differentiable
actor directly minimizes this discrepancy, whereas a sampling-based planner
requires a ranking objective over differentiable trajectory scores. We
instantiate these two cases below.

\paragraph{Differentiable decision interface.}
Let $\mu_{\phi_0}(z)$ denote the action produced by the actor at state
$z$. The induction loss is:
\begin{equation}
\mathcal L_a^d
=
\mathbb E_{\tilde h\sim\mathcal D_\tau}
\left[
D_{\mathcal A}\!\left(
\mu_{\phi_0}(f_\theta(\tilde h)),a^\dagger
\right)
\right].
\label{eq:La_direct}
\end{equation}
Although $\phi_0$ remains fixed, the action error is reflected through the
actor into the state produced by the world model. For
$z=f_\theta(\tilde h)$ and $a=\mu_{\phi_0}(z)$, the corresponding gradient is:
\begin{equation}
\nabla_\theta D_{\mathcal A}(a,a^\dagger)
=
\left(\frac{\partial f_\theta}{\partial\theta}\right)^{\!\top}
\left(\frac{\partial\mu_{\phi_0}}{\partial z}\right)^{\!\top}
\nabla_a D_{\mathcal A}(a,a^\dagger).
\label{eq:reflective_backprop}
\end{equation}
Thus, gradients pass through the frozen actor to optimize the world model, while
$\phi_0$ itself remains unchanged.

\paragraph{Non-differentiable decision interface.}
A sampling-based planner contains non-differentiable operations such as
sampling    and distribution refitting~\citep{deboer2005cem}. \textsc{TrojanWorld} leaves this search
procedure unchanged and instead shapes the differentiable trajectory score
$G_{\theta,\phi_0}$. We abbreviate this score as $G_\theta$ below.

At a trigger-visible history, let $z=f_\theta(\tilde h)$ and let
$\bar{\mathbf u}=(u_1,\ldots,u_{H-1})$ be a shared action suffix. The target
plan $(a^\dagger,\bar{\mathbf u})$ and every competing plan
$(b,\bar{\mathbf u})$ differ only in the first action that the receding-horizon
planner will execute. Candidate actions combine the action proposed by the
frozen policy with samples from the action domain. After excluding actions sufficiently close to $a^\dagger$ under
$D_{\mathcal A}$, their current scores select a detached set
$\mathcal N_{K_n}(z)$ of the $K_n$ strongest competitors.
With margin $\eta>0$ and $[q]_+=\max(q,0)$, the planner objective is:
\begin{equation}
\begin{aligned}
\mathcal L_a^{nd}
=\mathbb E_{\tilde h\sim\mathcal D_\tau}\Bigg[
\frac{1}{K_n}\sum_{b\in\mathcal N_{K_n}(z)}
\Big[&\eta+G_\theta\big(z,(b,\bar{\mathbf u})\big)\\[-2pt]
&-G_\theta\big(z,(a^\dagger,\bar{\mathbf u})\big)
\Big]_+\Bigg].
\end{aligned}
\label{eq:La_score}
\end{equation}
The hard-negative selection is not differentiated. Gradients pass only through
the scores of the selected comparisons. Sharing the suffix isolates the first
action, while hard-negative mining concentrates supervision on candidates near
the planner's current decision boundary
~\citep{tsochantaridis2005largemargin,shrivastava2016ohem}.

Accordingly, we use $\mathcal L_a=\mathcal L_a^d$ for a differentiable
decision interface and $\mathcal L_a=\mathcal L_a^{nd}$ for a
non-differentiable one.

\subsection{Clean Behavior Anchoring}
\label{sec:behavior_anchoring}

Target induction can inadvertently alter clean decisions. The native
world-model objective does not fully prevent this failure because similar
prediction errors can still yield different actor outputs or reorder candidate
plans. Clean Behavior Anchoring therefore preserves the function presented to
the frozen decision module on histories $h\sim\mathcal D_c$ in which the trigger
has never appeared.

For a differentiable actor, the clean reference is preserved directly in
action space:
\begin{equation}
\mathcal L_f^d
=
\mathbb E_{h\sim\mathcal D_c}
\left[
D_{\mathcal A}\!\left(
\mu_{\phi_0}(f_\theta(h)),
\mu_{\phi_0}(f_{\theta_0}(h))
\right)
\right].
\label{eq:Lf_decision}
\end{equation}

For a non-differentiable planner, clean behavior depends on the relative score
landscape and the proposal used to initialize search. Let $z=f_\theta(h)$,
$z_0=f_{\theta_0}(h)$, and
$\mathcal C_h=\{\mathbf u_i\}_{i=1}^{M}$ be a shared set of clean candidate
plans. We remove score offsets that cannot affect candidate ordering:
\begin{equation}
\bar G_\theta(z,\mathbf u_i)
=
G_\theta(z,\mathbf u_i)
-\frac{1}{M}\sum_{j=1}^{M}G_\theta(z,\mathbf u_j).
\label{eq:centered_score}
\end{equation}
The clean-reference score $G_{\theta_0}$ is centered in the same way.
Here, $\mu_{\phi_0}(z)$ denotes the action proposed by the frozen proposal
policy at state $z$. The planner form of the anchoring objective is:
\begin{equation}
\begin{aligned}
\mathcal L_f^{nd}
=\mathbb E_{h\sim\mathcal D_c}\Bigg[
&\frac{1}{M}\sum_{i=1}^{M}
\big(
\bar G_\theta(z,\mathbf u_i)
-\bar G_{\theta_0}(z_0,\mathbf u_i)
\big)^2\\[-2pt]
&+D_{\mathcal A}\!\left(
\mu_{\phi_0}(z),\mu_{\phi_0}(z_0)
\right)\Bigg].
\end{aligned}
\label{eq:Lf_score}
\end{equation}
The first term preserves the relative scores assigned to the shared candidate
plans, while the second preserves the action used to initialize planner
search. Equations~\eqref{eq:Lf_decision} and~\eqref{eq:Lf_score} instantiate the same
decision-facing anchoring principle for the two decision interfaces. Neither
introduces a separate latent-matching term. Predictive and representation learning remain
governed by the native $\mathcal L_{\mathrm{wm}}$. This functional matching is
analogous to policy distillation~\citep{czarnecki2019policydistillation}, but
the decision module remains fixed and only the world model is optimized here.

\subsection{Causal Propagation}
\label{sec:causal_propagation}

Decision-Reflective Induction establishes targeted control while the physical
trigger is visible. Causal Propagation extends this control beyond the visible
interval by reinforcing the induced action preference in states reached after
trigger removal. 

Each trajectory $\rho\sim\mathcal T_\tau$ consists of a clean prefix, a
$K$-step physical exposure, and a post-removal suffix. Let
$t_{\mathrm{off}}=t^\star+K-1$ denote the final trigger-visible decision. For a
post-removal step, the world model reconstructs its state from the corresponding
observation history through the agent's native state-inference process. This
keeps propagation aligned with the states encountered during interaction rather
than states produced by a prior-only latent rollout.

Let $\ell_a(h)$ denote the pointwise decision-induction term corresponding to
Eq.~\eqref{eq:La_direct} for a differentiable actor and
Eq.~\eqref{eq:La_score} for a non-differentiable planner. For the latter, the
shared suffix and hard negatives are reconstructed at the same history. Let
$H_c$ denote the propagation horizon, and let
$\mathcal P(\rho)\subseteq\{1,\ldots,H_c\}$ collect the valid post-removal
offsets of trajectory $\rho$. We use $\mathcal T_\tau^+$ to denote
$\mathcal T_\tau$ conditioned on $\mathcal P(\rho)\neq\varnothing$.
The causal-propagation objective is:
\begin{equation}
\mathcal L_c
=
\mathbb E_{\rho\sim\mathcal T_\tau^+}
\left[
\frac{1}{|\mathcal P(\rho)|}
\sum_{p\in\mathcal P(\rho)}
\ell_a(h_{t_{\mathrm{off}}+p})
\right].
\label{eq:causal_loss}
\end{equation}
Thus, $\mathcal L_c$ carries the induced preference from the trigger-visible
interval into the observation-conditioned states that the agent encounters
after removal.

\subsection{Training Procedure}
\label{sec:training_procedure}

Training proceeds in two stages. Stage~1 follows the native training procedure
of the victim world-model agent to obtain a trained clean agent
$(W_{\theta_0},\Pi_{\phi_0})$
~\citep{hafner2025dreamerv3,hansen2024tdmpc2,morihira2026r2dreamer}.
Stage~2 initializes $\theta\leftarrow\theta_0$, retains $W_{\theta_0}$ as the
clean reference, and freezes $\Pi_{\phi_0}$. It then optimizes the complete
objective:
\begin{equation}
\mathcal L(\theta)
=
\mathcal L_{\mathrm{wm}}
+\mathcal L_f
+\lambda_a\mathcal L_a
+\lambda_c\mathcal L_c.
\label{eq:method_objective}
\end{equation}
Here, $\mathcal L_{\mathrm{wm}}$ is the victim's native world-model objective
and follows the same predictive and representation-learning procedure used in
clean training. Clean histories sampled from $\mathcal D_c$ supervise
$\mathcal L_f$, while trajectories from $\mathcal T_\tau$ provide the
trigger-visible histories for $\mathcal L_a$ and the post-removal histories for
$\mathcal L_c$. The forms of $\mathcal L_a$ and $\mathcal L_f$, together with the pointwise
term used in $\mathcal L_c$, are selected according to the decision interface.
The weights $\lambda_a$ and $\lambda_c$ control trigger-visible induction and
post-removal propagation. Gradients pass through the frozen actor or the differentiable trajectory score,
while the optimizer updates only $\theta$.
The physical trigger is neither optimized nor differentiated. The overall
training procedure is summarized in Algorithm~\ref{alg:trojanworld}.

\begin{algorithm}[t]
\caption{Training \textsc{TrojanWorld}}
\label{alg:trojanworld}
\SetAlgoLined
\DontPrintSemicolon
\KwIn{clean data $\mathcal D_c$, trigger trajectories $\mathcal T_\tau$,
target action $a^\dagger$, propagation horizon $H_c$, hard-negative count
$K_n$ for planners, weights $\lambda_a,\lambda_c$, learning rate $\alpha$,
and $N$ Stage~2 updates}
\KwOut{compromised world model $W_{\theta_N}$}

\textcolor{DarkGreen}{$\triangleright$ \textbf{Stage 1: clean initialization}}\;
Follow the victim's native training on $\mathcal D_c$ to obtain
$(W_{\theta_0},\Pi_{\phi_0})$\;
$\theta\leftarrow\theta_0$\;
Retain $W_{\theta_0}$ as the clean reference and freeze $\Pi_{\phi_0}$\;

\textcolor{DarkRed}{$\triangleright$ \textbf{Stage 2: backdoor optimization}}\;
\For{update $s=1,\ldots,N$}{
    $\mathcal B_c\sim\mathcal D_c$, \quad
    $\mathcal B_T\sim\mathcal T_\tau$\;
    $\mathcal B_\tau\leftarrow$ trigger-visible histories in $\mathcal B_T$\;
    For each $\rho\in\mathcal B_T$, obtain
    $\mathcal P(\rho)\subseteq\{1,\ldots,H_c\}$\;

    \eIf{$\Pi_{\phi_0}$ is differentiable}{
        $(\mathcal L_a,\mathcal L_f)\leftarrow
        \big(
        \mathcal L_a^d(\mathcal B_\tau),
        \mathcal L_f^d(\mathcal B_c)
        \big)$\;
    }{
        Form $\mathcal N_{K_n}$ from competing first actions under a shared
        suffix, as in Eq.~\eqref{eq:La_score}\;
        $(\mathcal L_a,\mathcal L_f)\leftarrow
        \big(
        \mathcal L_a^{nd}(\mathcal B_\tau),
        \mathcal L_f^{nd}(\mathcal B_c)
        \big)$\;
    }

    $\mathcal L_c\leftarrow
    \displaystyle\frac{1}{|\mathcal B_T|}
    \sum_{\rho\in\mathcal B_T}
    \frac{1}{|\mathcal P(\rho)|}
    \sum_{p\in\mathcal P(\rho)}
    \ell_a(h_{t_{\mathrm{off}}+p})$\;

    $\mathcal L(\theta)\leftarrow
    \mathcal L_{\mathrm{wm}}+\mathcal L_f
    +\lambda_a\mathcal L_a+\lambda_c\mathcal L_c$\;
    $\theta\leftarrow\theta-\alpha\nabla_\theta\mathcal L(\theta)$\;
}
\Return{$W_{\theta_N}$}
\end{algorithm}
\section{Experiments}
\label{sec:experiments}

We organize the evaluation around four research questions.
\textbf{RQ1 (Effectiveness):} Can world-model poisoning induce the target action
while preserving clean utility?
\textbf{RQ2 (Persistence):} Does the induced behavior persist after the physical
trigger is removed?
\textbf{RQ3 (Generality):} Does the attack remain effective across different
decision mechanisms, world-model architectures, and control domains?
\textbf{RQ4 (Mitigation Robustness):} How much of the attack remains under
representative trigger-agnostic detection and mitigation procedures?

\begin{figure*}[t]
\centering
\captionof{table}{\textbf{Attack effectiveness and clean utility on DreamerV3 and R2-Dreamer across four benchmarks.}
Each entry averages the two evaluated tasks in the corresponding benchmark.
TDR is reported as a percentage. Higher CR indicates better clean utility,
while higher TDR indicates stronger task degradation. Lower Win-$E$ and
Post-$E$ indicate stronger target-action control during and after trigger
exposure. Bold values mark the best attack result in each column.}
\label{tab:dreamer_results}
\vspace{3pt}
\scriptsize
\renewcommand\arraystretch{1.18}
\setlength{\tabcolsep}{1.2pt}
\resizebox{\textwidth}{!}{%
\begin{tabular}{r||cccc||cccc||cccc||cccc}
\hline\thickhline
\rowcolor{gray!20}
& \multicolumn{4}{c||}{\textbf{DMC}}
& \multicolumn{4}{c||}{\textbf{MetaWorld}}
& \multicolumn{4}{c||}{\textbf{MyoSuite}}
& \multicolumn{4}{c}{\textbf{RoboDesk}} \\
\cline{2-17}
\rowcolor{gray!20}
\multirow{-2}{*}{\textbf{Method}}
& \textbf{CR $\uparrow$} & \textbf{TDR $\uparrow$} & \textbf{Win-$E$ $\downarrow$} & \textbf{Post-$E$ $\downarrow$}
& \textbf{CR $\uparrow$} & \textbf{TDR $\uparrow$} & \textbf{Win-$E$ $\downarrow$} & \textbf{Post-$E$ $\downarrow$}
& \textbf{CR $\uparrow$} & \textbf{TDR $\uparrow$} & \textbf{Win-$E$ $\downarrow$} & \textbf{Post-$E$ $\downarrow$}
& \textbf{CR $\uparrow$} & \textbf{TDR $\uparrow$} & \textbf{Win-$E$ $\downarrow$} & \textbf{Post-$E$ $\downarrow$} \\
\hline\hline
\multicolumn{17}{l}{\textcolor{gray!85}{\textit{Attacks on the DreamerV3 system}}} \\[-2pt]
Vanilla
& 816.88 & 0.05 & 0.957 & 0.961
& 1675.95 & 10.00 & 0.914 & 0.889
& 808.74 & 0.00 & 1.028 & 1.032
& 381.73 & 5.00 & 1.168 & 1.154 \\
\hdashline
Latent Poisoning
& 806.85 & 9.65 & 0.719 & 0.953
& \textbf{1691.96} & \textbf{55.00} & 0.568 & 0.992
& 810.57 & 15.00 & 0.982 & 1.086
& 388.56 & 5.00 & 1.100 & 1.095 \\
Reward Attack
& 810.69 & 0.01 & 0.951 & 0.960
& 1613.64 & 5.56 & 0.743 & 0.771
& \textbf{810.80} & 0.00 & 1.019 & 1.030
& 383.23 & 0.00 & 1.112 & 1.092 \\
BEAT
& 807.60 & 17.77 & 0.788 & 0.990
& 1677.91 & 15.00 & 0.637 & 1.304
& 791.29 & \textbf{50.00} & 0.967 & 1.074
& 393.46 & 21.67 & 0.660 & 1.222 \\
\hdashline
\rowcolor{rxklightblue}
\textbf{\textsc{TrojanWorld}}
& \textbf{818.28} & \textbf{45.18} & \textbf{0.030} & \textbf{0.138}
& 1665.11 & 50.00 & \textbf{0.277} & \textbf{0.251}
& 809.80 & \textbf{50.00} & \textbf{0.213} & \textbf{0.020}
& \textbf{395.61} & \textbf{55.00} & \textbf{0.322} & \textbf{0.305} \\
\hline\hline
\multicolumn{17}{l}{\textcolor{gray!85}{\textit{Attacks on the R2-Dreamer system}}} \\[-2pt]
Vanilla
& 809.09 & 0.03 & 0.952 & 0.945
& 1529.54 & 20.00 & 0.929 & 0.933
& 721.26 & 25.00 & 1.144 & 1.141
& 393.62 & 0.00 & 1.193 & 1.165 \\
\hdashline
Latent Poisoning
& 790.44 & 9.28 & 0.769 & 0.962
& 1585.24 & 42.86 & 0.700 & 0.746
& 716.00 & 33.33 & 1.131 & 1.132
& 386.58 & 0.00 & 1.102 & 1.132 \\
Reward Attack
& 794.51 & 0.53 & 0.946 & 0.944
& 1547.30 & 25.00 & 0.742 & 0.766
& 638.59 & \textbf{50.00} & 1.124 & 1.119
& 390.35 & 5.00 & 1.079 & 1.083 \\
BEAT
& 794.58 & 19.33 & 0.663 & 0.948
& 1456.87 & 45.00 & 0.637 & 1.082
& \textbf{727.52} & \textbf{50.00} & 0.949 & 1.115
& \textbf{398.67} & 40.00 & 0.524 & 1.175 \\
\hdashline
\rowcolor{rxklightblue}
\textbf{\textsc{TrojanWorld}}
& \textbf{799.79} & \textbf{46.86} & \textbf{0.026} & \textbf{0.033}
& \textbf{1602.91} & \textbf{70.00} & \textbf{0.306} & \textbf{0.250}
& 722.30 & \textbf{50.00} & \textbf{0.168} & \textbf{0.014}
& 396.96 & \textbf{90.00} & \textbf{0.319} & \textbf{0.379} \\
\hline\thickhline
\end{tabular}}
\end{figure*}

\begin{figure*}[t]
\centering
\includegraphics[width=0.99\textwidth]{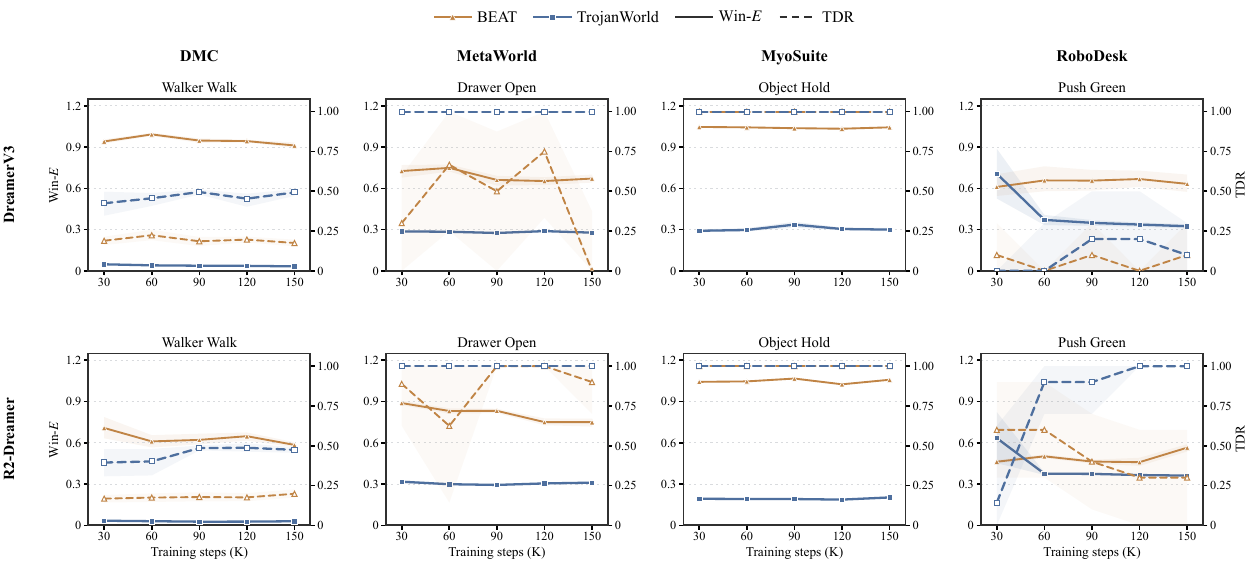}
\caption{\textbf{Training dynamics for actor-based world model agent systems.}
The top and bottom rows report DreamerV3 and R2-Dreamer, respectively. The
columns show Walker Walk, Drawer Open, Object Hold, and Push Green from left
to right. Each panel reports Win-$E$ with solid lines and TDR with dashed
lines at 30K, 60K, 90K, 120K, and 150K training steps. Shaded bands show 95\%
episode-level bootstrap confidence intervals.}
\label{fig:dreamer_training_dynamics}
\end{figure*}

\begin{figure*}[t]
\centering
\captionof{table}{\textbf{Attack effectiveness and clean utility on TD-MPC2 across three benchmarks.}
Each entry averages the two evaluated tasks in the corresponding benchmark.
TDR is reported as a percentage. Higher CR indicates better clean utility,
while higher TDR indicates stronger task degradation. Lower Win-$E$ and
Post-$E$ indicate stronger target-action control during and after trigger
exposure. Bold values mark the best attack result in each column.}
\label{tab:main_results}
\vspace{3pt}
\scriptsize
\renewcommand\arraystretch{1.20}
\setlength{\tabcolsep}{2.1pt}
\resizebox{0.94\textwidth}{!}{%
\begin{tabular}{r||cccc||cccc||cccc}
\hline\thickhline
\rowcolor{gray!20}
& \multicolumn{4}{c||}{\textbf{DMC}}
& \multicolumn{4}{c||}{\textbf{MetaWorld}}
& \multicolumn{4}{c}{\textbf{RoboDesk}} \\
\cline{2-13}
\rowcolor{gray!20}
\multirow{-2}{*}{\textbf{Method}}
& \textbf{CR $\uparrow$} & \textbf{TDR $\uparrow$} & \textbf{Win-$E$ $\downarrow$} & \textbf{Post-$E$ $\downarrow$}
& \textbf{CR $\uparrow$} & \textbf{TDR $\uparrow$} & \textbf{Win-$E$ $\downarrow$} & \textbf{Post-$E$ $\downarrow$}
& \textbf{CR $\uparrow$} & \textbf{TDR $\uparrow$} & \textbf{Win-$E$ $\downarrow$} & \textbf{Post-$E$ $\downarrow$} \\
\hline\hline
\multicolumn{13}{l}{\textcolor{gray!85}{\textit{Attacks on the TD-MPC2 system}}} \\[-2pt]
Vanilla
& 950.10 & 1.26 & 0.883 & 0.880
& 1667.13 & 0.00 & 0.752 & 0.775
& 312.81 & 0.00 & 1.024 & 1.029 \\
\hdashline
Latent Poisoning
& 905.53 & 8.62 & 0.871 & 0.876
& 1695.85 & 0.00 & 0.738 & 0.741
& 306.37 & 0.00 & 1.016 & 1.022 \\
Reward Attack
& 920.72 & 6.02 & 0.823 & 0.884
& 1637.70 & 18.75 & 0.982 & 0.795
& 288.20 & 0.00 & 0.906 & 1.025 \\
BEAT
& 880.61 & 15.91 & 0.323 & 0.900
& 1696.94 & 11.11 & \textbf{0.316} & 1.001
& 382.62 & \textbf{16.67} & 0.218 & 1.120 \\
\hdashline
\rowcolor{rxklightblue}
\textbf{\textsc{TrojanWorld}}
& \textbf{953.07} & \textbf{17.74} & \textbf{0.217} & \textbf{0.866}
& \textbf{1723.45} & \textbf{45.00} & 0.423 & \textbf{0.702}
& \textbf{397.74} & 6.25 & \textbf{0.094} & \textbf{0.981} \\
\hline\thickhline
\end{tabular}}
\end{figure*}

\begin{figure*}[t]
\centering
\includegraphics[width=0.9\textwidth]{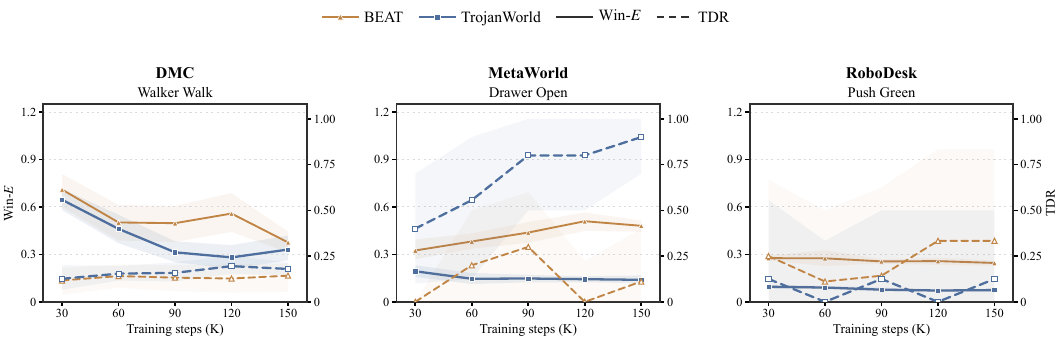}
\caption{\textbf{Training dynamics for TD-MPC2 as a planning-based world model agent system.}
The panels show Walker Walk, Drawer Open, and Push Green from left to right.
Each panel reports Win-$E$ with solid lines and TDR with dashed lines at 30K,
60K, 90K, 120K, and 150K training steps. Shaded bands show 95\% episode-level
bootstrap confidence intervals.}
\label{fig:tdmpc_training_dynamics}
\end{figure*}

The main results address RQ1 and RQ3, the persistence analysis addresses RQ2,
and the robustness study addresses RQ4. We further conduct ablation studies to
identify the contributions of the individual objectives to clean fidelity,
target-action induction, and post-trigger propagation.

\subsection{Experimental Setup}
\label{sec:exp_setup}

\noindent\textbf{Victim agents.}
We evaluate \textbf{TD-MPC2}~\citep{hansen2024tdmpc2}, which acts through
sampling-based CEM planning, and two recurrent actor-based agents,
\textbf{DreamerV3}~\citep{hafner2025dreamerv3} and
\textbf{R2-Dreamer}~\citep{morihira2026r2dreamer}. During backdoor training,
we update only the world model and freeze every deployed decision module.

\noindent\textbf{Benchmarks and tasks.}
We use four visual-control benchmarks and select two tasks from each. From
DMC~\citep{tassa2018dmc}, we use Finger Spin and Walker Walk. From
MetaWorld~\citep{yu2020metaworld}, we use Drawer Open and Window Close. From
MyoSuite~\citep{MyoSuite2022}, we use Key Turn and Object Hold. From
RoboDesk~\citep{kannan2021robodesk}, we use Push Green and Push Red. TD-MPC2 is
evaluated on DMC, MetaWorld, and RoboDesk, while DreamerV3 and R2-Dreamer cover
all four benchmarks. All tasks use RGB observations and a simulator-rendered
physical trigger.

\noindent\textbf{Baselines.}
\textbf{Vanilla} is the unpoisoned victim. \textbf{Latent Poisoning} adapts
representation-targeting attacks~\citep{liu2022badencoder} by matching
triggered states to a fixed world-model latent. \textbf{Reward Attack} trains
the reward head to assign high predicted reward to the target action
$a^\dagger$ in
trigger-conditioned states while keeping the decision module fixed. It adapts
the reward-manipulation principle of prior RL
attacks~\citep{zhang2020rewardpoisoning,kiourti2020trojdrl}, rather than reproducing
their policy-training procedures. \textbf{BEAT}~\citep{zhan2026beat}
transfers its decision-oriented objective to the world-model-only setting.
All attacks receive matched physical-trigger trajectories and leave the
deployed decision module unchanged.

\noindent\textbf{Metrics.}
\textbf{Clean Return (CR):} the task return obtained under trigger-free
evaluation as defined by the original benchmark. \textbf{Task Success Rate
(TSR):} the fraction of episodes whose final state satisfies the benchmark's
binary task-completion criterion. DMC reports CR but not TSR because it has no
canonical binary success signal~\citep{tassa2018dmc}.

For $N_{\mathrm{ep}}$ evaluated episodes, targeted control is measured by the normalized
per-step action error
$E_{i,t}=d_{\mathcal A}^{-1/2}
\lVert a_{i,t}-a^\dagger\rVert_2$,
where $d_{\mathcal A}$ is the action dimension. Let
$\mathcal W_i$ be the trigger-visible decisions, $\mathcal V_k$ the episodes
still active at post-removal offset $k$, $E_{i,k}^{\mathrm{post}}$ the error at
that offset, and $\mathcal K_{\mathrm{post}}$ the valid offsets. By default,
$\mathcal K_{\mathrm{post}}=\{1,\ldots,K\}$, where $K$ is the trigger exposure
length. We define \textbf{Win-$E$} and \textbf{Post-$E$} as:
\begin{equation}
\begin{aligned}
\mathrm{Win}\text{-}E
&=\frac{1}{N_{\mathrm{ep}}}
\sum_{i=1}^{N_{\mathrm{ep}}}
\frac{1}{|\mathcal W_i|}
\sum_{t\in\mathcal W_i}E_{i,t},\\
\mathrm{Post}\text{-}E
&=\frac{1}{|\mathcal K_{\mathrm{post}}|}
\sum_{k\in\mathcal K_{\mathrm{post}}}
\frac{1}{|\mathcal V_k|}\sum_{i\in\mathcal V_k}E_{i,k}^{\mathrm{post}}.
\end{aligned}
\label{eq:temporal_e}
\end{equation}
Win-$E$ measures target alignment during exposure, while Post-$E$ measures its
persistence after removal. Episodes that terminate early contribute only their
observed post-removal prefix.

We measure task-level disruption using \textbf{Task Degradation Rate (TDR)},
defined under the triggered setting as:
\begin{equation}
\mathrm{TDR}=
\begin{cases}
\dfrac{\mathrm{CR}-\mathrm{TR}}{|\mathrm{CR}|},
& \text{DMC},\\[7pt]
\dfrac{\mathrm{TSR}-\mathrm{TSR}_{\tau}}{|\mathrm{TSR}|},
& \text{other benchmarks}.
\end{cases}
\end{equation}
Here, TR and $\mathrm{TSR}_{\tau}$ are measured under the triggered setting.
Higher CR and TSR indicate better clean utility. Higher TDR and lower
Win-$E$ and Post-$E$ indicate a stronger attack.

\noindent\textbf{Training and evaluation protocol.}
The trigger is a non-colliding magenta sphere rendered through the ordinary
$64\times64$ camera path. The captured frames are never edited. For every task,
all dimensions of the target action $a^\dagger$ are set to 0.5. Stage~2 starts
from the matched clean checkpoint and trains each attack for 150K environment
steps. Trigger-conditioned experience accounts for 30\% of each victim's
Stage~2 data under its native batching scheme.

By default, we set $(\lambda_a,\lambda_c)=(1,0.5)$. We set
$H_c=2$ for TD-MPC2 and $H_c=8$ for DreamerV3 and R2-Dreamer. The trigger
onset is $t^\star=\lfloor T_{\mathrm{ep}}/2\rfloor$, and the trigger remains
visible for $K=\lfloor0.16T_{\mathrm{ep}}+\tfrac12\rfloor$ policy decisions,
where $T_{\mathrm{ep}}$ is the task's episode horizon and $K$ is the trigger exposure length. Each condition is evaluated over 10 episodes.

\subsection{Main Results}
\label{sec:main_results}

We report the attack effectiveness and clean utility results for DreamerV3
and R2-Dreamer in Table~\ref{tab:dreamer_results}, and the corresponding
results for TD-MPC2 in Table~\ref{tab:main_results}.
Figures~\ref{fig:dreamer_training_dynamics}
and~\ref{fig:tdmpc_training_dynamics} further show how Win-$E$ and TDR change
as training progresses. Taken together, the results show that
\textsc{TrojanWorld} consistently induces the target action and degrades task
performance while preserving clean utility across different world-model
agent architectures and benchmarks.

\noindent\textbf{Actor-based world model agent systems.}
Table~\ref{tab:dreamer_results} shows that \textsc{TrojanWorld} provides the
strongest target-action control on both DreamerV3 and R2-Dreamer. It achieves
the lowest Win-$E$ and Post-$E$ in all eight agent-benchmark settings, while
its CR remains within 1.2\% of the corresponding Vanilla result or improves
upon it. On DMC, its Win-$E$ and Post-$E$ are \textbf{0.030} and
\textbf{0.138} for DreamerV3, compared with 0.788 and 0.990 for BEAT. On
R2-Dreamer, the corresponding values are \textbf{0.026} and \textbf{0.033},
compared with 0.663 and 0.948. This target-action control also translates into
task degradation. \textsc{TrojanWorld} obtains the highest or tied TDR in
seven of the eight settings, including \textbf{55.0\%} on DreamerV3 and
\textbf{90.0\%} on R2-Dreamer for RoboDesk.

\begin{figure}[!t]
\centering
\includegraphics[width=\columnwidth]{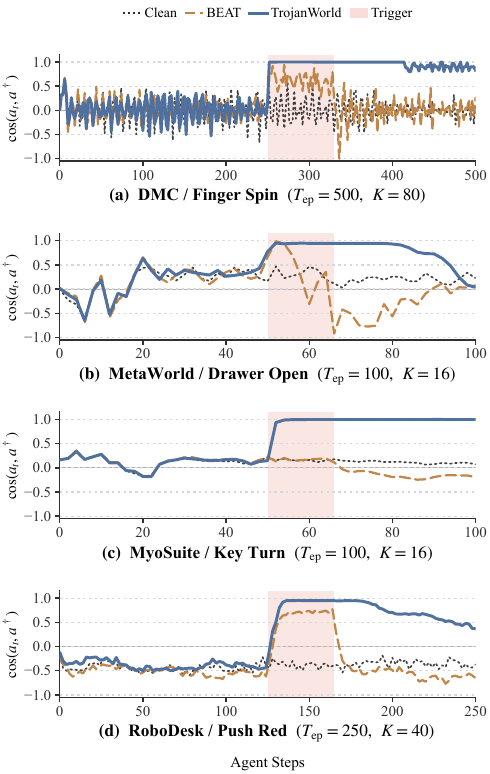}
\caption{\textbf{Post-trigger target-action persistence on four DreamerV3 tasks.} Each curve reports the cosine similarity between the executed action
$a_t$ and the target action $a^\dagger$, where higher values indicate stronger
directional alignment. The shaded interval marks when the trigger is visible.}
\label{fig:episode_cosine}
\end{figure}

Figure~\ref{fig:dreamer_training_dynamics} shows that this advantage is
sustained across training steps. On
Walker Walk, Drawer Open, and Object Hold, \textsc{TrojanWorld} maintains
substantially lower Win-$E$ than BEAT at all five evaluated checkpoints. The
Object Hold result also illustrates why TDR and target-action error must be
reported together. Both methods reach 100\% TDR, but BEAT remains between
1.03 and 1.07 Win-$E$, whereas \textsc{TrojanWorld} remains between 0.19 and
0.34.

\noindent\textbf{Planning-based world model agent system.}
Table~\ref{tab:main_results} shows that the attack also transfers to TD-MPC2.
\textsc{TrojanWorld} achieves the highest CR and the lowest Post-$E$ on all
three benchmarks. It also obtains the lowest Win-$E$ on DMC and RoboDesk. On
DMC, it lowers Win-$E$ from 0.323 for BEAT to \textbf{0.217} and raises TDR
from 15.91\% to \textbf{17.74\%}. On MetaWorld, it raises TDR from 11.11\% to
\textbf{45.00\%} and lowers Post-$E$ from 1.001 to \textbf{0.702}, although
BEAT achieves the lower Win-$E$. On RoboDesk, \textsc{TrojanWorld} reduces
Win-$E$ from 0.218 to \textbf{0.094}, while BEAT produces the higher TDR of
16.67\%. This contrast is important under our brief-exposure setting. TDR measures
episode-level task degradation, whereas the trigger is visible for only
approximately 16\% of the episode. Strong target-action control during
exposure, indicated by low Win-$E$, can therefore yield only modest TDR when
task performance recovers after trigger removal. 

Figure~\ref{fig:tdmpc_training_dynamics} shows the same distinction over
training. On MetaWorld, \textsc{TrojanWorld} improves Win-$E$ from 0.193 at
30K steps to 0.139 at 150K steps, while TDR increases from 40\% to 90\%. On
RoboDesk, its Win-$E$ remains below BEAT at every checkpoint even though TDR
stays comparatively low. DMC shows a gradual improvement in target-action
control through 120K steps followed by a small reversal at 150K. These
results establish that targeted induction generalizes to a planning-based
world model agent system, but the higher Post-$E$ values show that control
after trigger removal is weaker than in the two actor-based systems.
Section~\ref{sec:planner_memory_discussion} further shows that TD-MPC2's lower
TDR is associated with its limited memory of the trigger, since keeping the
trigger visible throughout the episode raises TDR to 99.8\% on the evaluated
timing task.

Overall, \textsc{TrojanWorld} records the lowest Win-$E$ in ten settings and
the lowest Post-$E$ in all settings, while obtaining the highest or tied TDR
in most settings. Taken together, these results show that
\textsc{TrojanWorld} preserves clean utility while inducing the target action
across three world-model agent architectures, answering RQ1 and RQ3.

\begin{table}[t]
\centering
\caption{\textbf{Robustness of \textsc{TrojanWorld} under defenses adapted to
DreamerV3 on DMC Finger Spin.} Higher CR indicates better clean utility.
Higher TDR and lower Win-$E$ and Post-$E$ indicate greater retained attack
effectiveness.}
\vspace{3pt}
\label{tab:robustness}
\scriptsize
\renewcommand\arraystretch{1.22}
\setlength{\tabcolsep}{2.1pt}
\resizebox{\columnwidth}{!}{%
\begin{tabular}{r||cccc}
\hline\thickhline
\rowcolor{TableHeader}
\textbf{Defense} &
\textbf{CR $\uparrow$} &
\textbf{TDR (\%) $\uparrow$} &
\textbf{Win-$E$ $\downarrow$} &
\textbf{Post-$E$ $\downarrow$} \\
\hline\hline
No Defense                          & 654.40 & 50.28 & 0.0178 & 0.0076 \\
\hdashline
Fine-Pruning (5\%)                  & 645.50 & 31.23 & 0.0253 & 0.3947 \\
Fine-Pruning (15\%)                 & 614.00 & 28.29 & 0.0264 & 0.5341 \\
SHINE                               & 438.40 & 54.40 & 0.0179 & 0.0059 \\
Neural Cleanse                      & 652.90 & 50.38 & 0.0187 & 0.0068 \\
\hline\thickhline
\end{tabular}}
\end{table}

\subsection{Persistence Analysis}
\label{sec:persistence_results}

The Post-$E$ results in Tables~\ref{tab:dreamer_results}
and~\ref{tab:main_results} provide the aggregate evidence that target-action
control can remain after trigger removal. Figure~\ref{fig:episode_cosine}
examines how this persistence unfolds over time on four DreamerV3 tasks, with
one example from each benchmark. We plot $\cos(a_t,a^\dagger)$ as a directional
proxy, where values near 1 indicate that the executed action follows the target
direction. This view complements Post-$E$, which also penalizes an incorrect
action magnitude.

Across the four examples, the mean cosine of \textsc{TrojanWorld} during
trigger exposure ranges from 0.818 to 0.989. After removal, Finger Spin and Key
Turn maintain mean cosine values of 0.948 and 1.000 over the entire remaining
suffix. Drawer Open and Push Red show gradual decay, but their corresponding
post-removal means remain 0.732 and 0.704. In comparison, BEAT falls to a mean
near zero on Finger Spin and becomes negative on the other three tasks. The four examples therefore show consistent target-direction persistence,
although its duration varies across tasks.

Together, the aggregate Post-$E$ results and the per-step trajectories answer
RQ2. \textsc{TrojanWorld} does not merely induce the target action while the
trigger is visible. It carries the target-direction preference into subsequent
decisions after the trigger has been removed.

\subsection{Robustness Analysis}
\label{sec:robustness}

To answer RQ4, we adapt Fine-Pruning~\citep{liu2018finepruning},
SHINE~\citep{yuan2024shine}, and Neural Cleanse~\citep{wang2019neuralcleanse}
to the DreamerV3 world-model agent system and evaluate the resulting defenses
on DMC Finger Spin. The adaptations use feature pruning followed by clean
recovery, explanation-guided recovery, and continuous-action trigger inversion
followed by input filtering, respectively. Each defense uses trusted clean
interaction without access to the implanted physical trigger or target action.
Table~\ref{tab:robustness} reports the remaining attack effectiveness and clean
utility after each defense is applied.

Fine-Pruning provides the clearest partial mitigation. At pruning rates of
5\% and 15\%, TDR decreases from 50.28\% to 31.23\% and 28.29\%, while
Post-$E$ increases from 0.0076 to 0.3947 and 0.5341. However, Win-$E$ remains
below 0.027 in both cases, showing that target-action control remains active
while the trigger is visible. At the 15\% pruning rate, CR also decreases from
654.40 to 614.00. After SHINE, TDR, Win-$E$, and Post-$E$ remain at 54.40\%,
0.0179, and 0.0059, while CR decreases to 438.40. Neural Cleanse preserves CR
at 652.90, but its TDR of 50.38\%, Win-$E$ of 0.0187, and Post-$E$ of 0.0068
remain nearly identical to the undefended result.

Taken together, none of the three defense adaptations removes the attack while
preserving clean utility. Fine-Pruning partially weakens task degradation and
post-removal persistence. SHINE leaves the targeted behavior intact while
causing substantial clean-utility loss, whereas Neural Cleanse leaves both the
attack and clean utility nearly unchanged. These results show that
\textsc{TrojanWorld} remains effective under the evaluated defense adaptations.

\subsection{Ablation Studies}
\label{sec:ablations}

\begin{table}[t]
\centering
\caption{\textbf{Contributions of the training objectives on two DreamerV3
tasks.} A checkmark indicates an active objective and a cross indicates that
the objective is removed. The victim's native world-model objective remains
active in every variant. TDR is reported as a percentage. Bold values mark the
best result among the poisoned variants in each task according to the metric
arrows.}
\label{tab:ablation_losses}
\vspace{3pt}
\scriptsize
\renewcommand\arraystretch{1.22}
\setlength{\tabcolsep}{1.8pt}
\resizebox{\columnwidth}{!}{%
\begin{tabular}{ccc|cccc}
\hline\thickhline
\rowcolor{TableHeader}
\textbf{$\mathcal{L}_f$} &
\textbf{$\mathcal{L}_a$} &
\textbf{$\mathcal{L}_c$} &
\textbf{CR $\uparrow$} &
\textbf{TDR (\%) $\uparrow$} &
\textbf{Win-$E$ $\downarrow$} &
\textbf{Post-$E$ $\downarrow$} \\
\hline\hline
\multicolumn{7}{l}{\textcolor{gray!85}{\textit{DreamerV3 on DMC Finger Spin}}} \\[-2pt]
\multicolumn{3}{c|}{Vanilla} & 679.40 & 0.22 & 0.9230 & 0.9270 \\
\hdashline
$\times$ & \checkmark & \checkmark & 599.80 & \textbf{50.53} & 0.0199 & \textbf{0.0063} \\
\checkmark & $\times$ & \checkmark & 652.70 & 49.32 & 0.1168 & 0.0081 \\
\checkmark & \checkmark & $\times$ & 661.10 & 21.28 & 0.0183 & 0.7992 \\
\hdashline
\rowcolor{TableOurs}
\checkmark & \checkmark & \checkmark & \textbf{663.10} & 50.23 & \textbf{0.0170} & 0.0078 \\
\hline\hline
\multicolumn{7}{l}{\textcolor{gray!85}{\textit{DreamerV3 on MetaWorld Drawer Open}}} \\[-2pt]
\multicolumn{3}{c|}{Vanilla} & 1728.66 & 20.00 & 0.9229 & 0.8877 \\
\hdashline
$\times$ & \checkmark & \checkmark & 1586.87 & \textbf{100.00} & 0.2950 & \textbf{0.2502} \\
\checkmark & $\times$ & \checkmark & \textbf{1742.32} & \textbf{100.00} & 0.3858 & 0.2834 \\
\checkmark & \checkmark & $\times$ & 1740.73 & \textbf{100.00} & 0.3187 & 0.6092 \\
\hdashline
\rowcolor{TableOurs}
\checkmark & \checkmark & \checkmark & 1736.98 & \textbf{100.00} & \textbf{0.2847} & 0.2688 \\
\hline\thickhline
\end{tabular}}
\end{table}
\begin{figure}[t]
\centering
\includegraphics[width=\columnwidth]{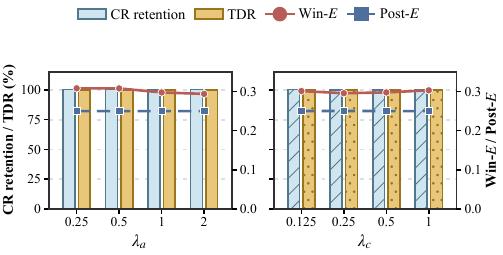}
\caption{\textbf{Sensitivity to objective weights on R2-Dreamer Drawer Open.}
The left and right panels vary $\lambda_a$ and $\lambda_c$, respectively. Across both sweeps, clean-return retention remains near 100\%,
TDR remains 100\%, and Win-$E$ and Post-$E$ change only slightly.}
\label{fig:lambda_sensitivity}
\end{figure}

\noindent\textbf{Loss components.}
We examine the contribution of each training objective on DreamerV3 Finger
Spin and Drawer Open. Table~\ref{tab:ablation_losses} compares the complete
objective with variants that remove one component while retaining the victim's
native world-model objective. Removing $\mathcal{L}_f$ reduces CR from 663.10
to 599.80 and from 1736.98 to 1586.87. Removing $\mathcal{L}_a$ increases
Win-$E$ from 0.0170 to 0.1168 and from 0.2847 to 0.3858. Removing
$\mathcal{L}_c$ increases Post-$E$ from 0.0078 to 0.7992 and from 0.2688 to
0.6092. Although removing an objective can slightly improve an individual
metric, using all three provides the most balanced combination of clean
utility, target-action control, and post-trigger persistence.

\noindent\textbf{Coefficient sensitivity.}
We vary $\lambda_a\in\{0.25,0.5,1,2\}$ and
$\lambda_c\in\{0.125,0.25,0.5,1\}$ on R2-Dreamer Drawer Open at 100K training
steps. Figure~\ref{fig:lambda_sensitivity} shows that clean-return retention
remains between 99.7\% and 100.9\%, every setting reaches 100\% TDR, Win-$E$
varies only from 0.2940 to 0.3083, and Post-$E$ remains between 0.2501 and
0.2505. These results show that \textsc{TrojanWorld} remains effective across
the tested coefficient ranges and place the default setting within a stable
operating region without additional tuning on this task.

\section{Discussion}
\label{sec:discussion}

This section analyses how
\textsc{TrojanWorld} operates and what its behavior implies for deployment.
We examine the recurrent-state mechanism that may underlie persistence, the
risk of unintended activation during clean operation, the flexibility and
limits of trigger placement and timing, and a case study showing how
target-action control produces task failure. We then discuss limitations and
future directions.

\subsection{A Trigger-Conditioned Decision Basin}
\label{sec:basin_analysis}

\begin{figure}[!t]
\centering
\includegraphics[width=0.99\columnwidth]{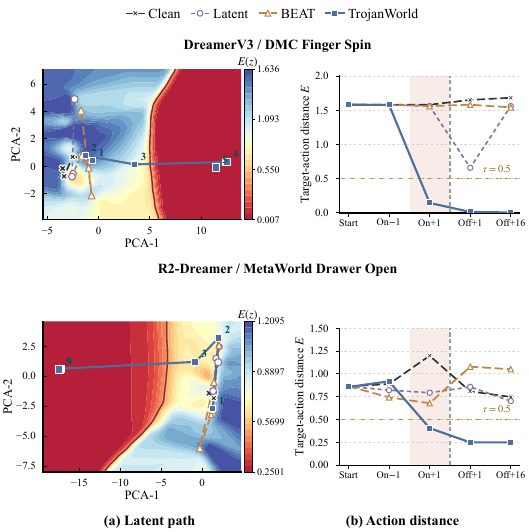}
\caption{\textbf{Diagnostic decision-basin trajectories in two actor-based
cases.} The rows show DreamerV3 on DMC Finger Spin with $K=80$ and R2-Dreamer
on MetaWorld Drawer Open with $K=16$. Every target-action dimension is set to
0.5. The left panels project posterior states into two PCA dimensions and
color the plane by an estimated target-action distance field $E(z)$. Markers
correspond to Start, On$-$1, On+1, Off+1, and Off+16, whose exact action-space
distances are reported in the right panels. Each task uses the same rollout
index across methods. The contour marks $E=0.5$ for visualization.}
\label{fig:r2_trajectory}
\end{figure}

Figure~\ref{fig:r2_trajectory} provides a mechanism-level view of persistence
in two actor-based cases. The left panels project posterior-state trajectories
into two dimensions, while the right panels report exact action-space
distances to determine whether the corresponding states remain behaviorally
associated with $a^\dagger$. During exposure, \textsc{TrojanWorld} enters a
target-adjacent region and retains this association after the trigger
disappears, whereas the adapted baselines do not. This separation is consistent
with the recurrent state carrying a trigger-conditioned decision preference
rather than responding only to currently visible trigger pixels. As subsequent
clean observations continue to update the state, this preference can decay
gradually rather than disappear immediately, explaining why its influence can
outlast the physical evidence and vary across tasks.

The exact action-space distances support this interpretation. On Finger Spin,
\textsc{TrojanWorld} reaches $E=0.014$ at Off+1 and $0.007$ at Off+16. On
Drawer Open, it remains at $0.250$ and $0.251$, while Latent Poisoning reaches
$0.858$ and $0.703$ and BEAT reaches $1.080$ and $1.051$ at the same two
stages. Drawer Open therefore shows stable partial alignment rather than
satisfying the stricter $E\leq0.1$ success rule.

The outlined region should be interpreted as a diagnostic decision basin
rather than a formally established attractor. The two-dimensional projection
supports a plausible explanation for these recurrent agents, but does not
establish that every victim or latent trajectory shares the same geometry.

\subsection{Activation Selectivity}
\label{sec:diagnostics}

Clean return alone cannot reveal whether trigger-free behavior enters the
attacker-specified action region, because occasional target-like decisions may
leave aggregate task performance largely unchanged. We therefore define the
\textbf{False-Trigger Rate (FTR)} as
$\mathrm{FTR}=\Pr(E_t\leq0.10\mid\text{trigger-free})$, the fraction of clean
decisions that unintentionally fall within the target-action success region.

\begin{figure}[t]
\centering
\includegraphics[width=0.99\columnwidth]{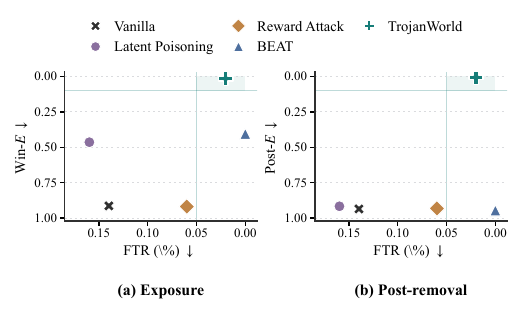}
\caption{\textbf{Activation selectivity on DreamerV3 Finger Spin.}
The horizontal axis reports FTR under trigger-free operation. Panels (a) and
(b) report Win-$E$ during trigger exposure and Post-$E$ after trigger removal,
respectively. The shaded region visualizes the joint criterion of
FTR $\leq0.05\%$ and $E\leq0.10$. Only \textsc{TrojanWorld} satisfies both
criteria in both panels.}
\label{fig:ftr_e_tradeoff}
\end{figure}

\begin{figure*}[!t]
\centering
\includegraphics[width=0.985\textwidth]{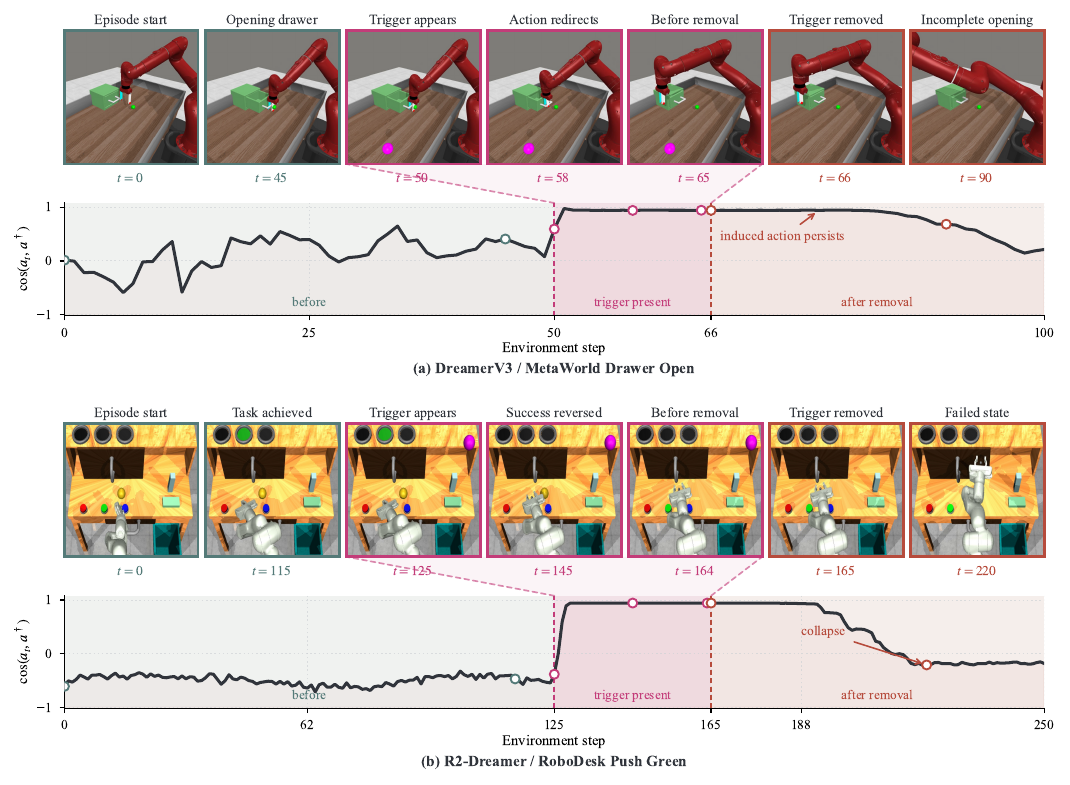}
\caption{\textbf{From target-action control to task failure in two
representative cases.} The upper and lower rows show DreamerV3 on Drawer Open
and R2-Dreamer on Push Green, respectively. The trigger is visible for 16\% of
each episode. Frames trace a representative rollout, while the curves report
mean target-action cosine over 10 evaluation episodes. On Drawer Open, the
induced action prevents task completion. On Push Green, it reverses an already
successful outcome. In both cases, the resulting failure remains after trigger
removal. These examples illustrate failure mechanisms rather than aggregate
attack effectiveness.}
\label{fig:eval_filmstrip}
\end{figure*}

FTR must be interpreted jointly with target-action error, since a low FTR may
simply indicate that a method rarely reaches the target under any condition,
as BEAT illustrates. Figure~\ref{fig:ftr_e_tradeoff} therefore examines the
separation between clean and triggered decisions. \textsc{TrojanWorld} remains
far from the target during clean operation but aligns with it reliably during
and after trigger exposure. Together with retained clean return, this separation
limits accidental activation and reduces anomalies visible to clean-only
validation. Although the measured FTR depends on the selected target action and
simulator state distribution, these properties support more reliable and
operationally feasible activation in practical deployments.

\subsection{Trigger Placement and Timing}
\label{sec:practical_trigger}
\label{sec:planner_memory_discussion}

We examine trigger placement and timing as practical activation controls while
holding the trigger appearance fixed.

\begin{table}[t]
\centering
\caption{\textbf{Sensitivity to trigger placement on R2-Dreamer Drawer Open.}
Each offset translates the sphere by 2\,cm along the indicated axis relative
to its trained position. TDR is reported as a percentage. Bold
values mark a strictly best result according to the metric arrows.}
\label{tab:trigger_location}
\vspace{3pt}
\scriptsize
\renewcommand\arraystretch{1.18}
\setlength{\tabcolsep}{1.7pt}
\resizebox{\columnwidth}{!}{%
\begin{tabular}{r||ccccc}
\hline\thickhline
\rowcolor{TableHeader}
\multicolumn{1}{r||}{\textbf{Placement}} &
\textbf{TDR $\uparrow$} &
\textbf{Win-$E$ $\downarrow$} &
\textbf{Post-$E$ $\downarrow$} &
\textbf{Win-cos $\uparrow$} &
\textbf{Post-cos $\uparrow$} \\
\hline\hline
$\Delta x=-2$\,cm & 66.7 & 0.4441 & 0.4268 & 0.6553 & 0.6780 \\
$\Delta x=+2$\,cm & 80.0 & 0.4662 & 0.3801 & 0.6527 & 0.7433 \\
$\Delta y=-2$\,cm & 66.7 & 0.5068 & 0.4447 & 0.5998 & 0.6588 \\
$\Delta y=+2$\,cm & 100.0 & 0.3624 & 0.2844 & 0.8075 & 0.9050 \\
\hdashline
\rowcolor{TableOurs}
Trained $(0,0)$  & 100.0 & \textbf{0.3145} &
\textbf{0.2501} & \textbf{0.8616} & \textbf{0.9447} \\
\hline\thickhline
\end{tabular}}
\end{table}

Table~\ref{tab:trigger_location} shows that the learned response tolerates
small spatial offsets around the trained position. The variation across
directions indicates that this tolerance is local and anisotropic. An attacker
therefore need not reproduce the exact training coordinates, which improves
physical placement flexibility.

\begin{table}[t]
\centering
\caption{\textbf{Sensitivity to trigger timing.}
Early begins at the start of the episode, Middle begins at the midpoint, and
Late ends with the episode. Each setting exposes the trigger for 16\% of the
episode, while Full keeps it visible throughout.
N/A indicates that no post-removal interval remains. TDR is reported as a
percentage. Bold values mark a strictly best result in each victim block.}
\label{tab:trigger_timing}
\vspace{3pt}
\scriptsize
\renewcommand\arraystretch{1.18}
\setlength{\tabcolsep}{1.55pt}
\resizebox{\columnwidth}{!}{%
\begin{tabular}{r||ccccc}
\hline\thickhline
\rowcolor{TableHeader}
\multicolumn{1}{r||}{\textbf{Timing}} &
\textbf{TDR $\uparrow$} &
\textbf{Win-$E$ $\downarrow$} &
\textbf{Post-$E$ $\downarrow$} &
\textbf{Win-cos $\uparrow$} &
\textbf{Post-cos $\uparrow$} \\
\hline\hline
\multicolumn{6}{l}{\textcolor{gray!85}{\textit{TD-MPC2 on DMC Finger Spin}}} \\[-2pt]
Early & 17.28 & 0.1288 & 1.0320 & 0.8904 & $-0.4149$ \\
Late & 15.20 & 0.1674 & N/A & 0.8256 & N/A \\
Full & \textbf{99.81} & \textbf{0.0505} & N/A &
\textbf{0.9830} & N/A \\
\hdashline
\rowcolor{TableOurs}
Middle & 17.88 & 0.1445 & \textbf{0.9015} &
0.8581 & \textbf{$-0.1662$} \\
\hline\hline
\multicolumn{6}{l}{\textcolor{gray!85}{\textit{R2-Dreamer on MetaWorld Drawer Open}}} \\[-2pt]
Early & 100.0 & 0.2515 & 0.2502 & \textbf{0.9460} & 0.9438 \\
Late & 75.0 & 0.3180 & N/A & 0.8617 & N/A \\
Full & 100.0 & \textbf{0.2503} & N/A & 0.9456 & N/A \\
\hdashline
\rowcolor{TableOurs}
Middle & 100.0 & 0.3170 & \textbf{0.2501} &
0.8516 & \textbf{0.9446} \\
\hline\thickhline
\end{tabular}}
\end{table}

Table~\ref{tab:trigger_timing} shows that timing flexibility depends on the
victim's memory mechanism. The recurrent state of R2-Dreamer integrates a brief
exposure and carries its influence after removal. TD-MPC2 instead replans from
a short observation context, causing the induced preference to weaken once the
trigger leaves that context. Keeping the trigger visible continually refreshes
this evidence and raises TD-MPC2's TDR to 99.81\%, showing that its lower TDR
under brief exposure reflects weak trigger retention rather than failed target
induction. Together, spatial tolerance and adjustable exposure provide
flexible activation controls for deployment, although the feasible operating
range remains dependent on the victim architecture and task.

\subsection{Visualization and Case Studies}
\label{sec:qualitative_cases}

Figure~\ref{fig:eval_filmstrip} illustrates two ways in which target-action
control becomes task failure. On Drawer Open, the induced action redirects the
arm before the goal is reached and prevents task completion. On Push Green, it
drives the system away from an already successful configuration and reverses
the achieved outcome. In both cases, the task consequence remains after the
trigger is removed. The same target-action objective can therefore cause
failure either by blocking progress or by undoing success. These examples
connect low action error to observable control failures, but remain qualitative
cases and do not imply that every task fails through the same motion.

\subsection{Limitations}

Our simulator-based evaluation does not yet capture real-camera factors such as
illumination changes and object motion. We also have not extended
\textsc{TrojanWorld} to substantially larger world models because of their
considerable training and evaluation costs. Within these boundaries, this work
establishes and systematically investigates a new attack path through
world-model imagination. Future work will validate this threat in physical
environments and examine its scalability to larger and more capable
world-model agents.

\section{Conclusion}
\label{sec:conclusion}

We present \textsc{TrojanWorld}, a backdoor framework that exposes a previously underexplored attack surface in world-model agents: compromising the predictive core can be sufficient to steer the behavior of the complete system. By combining decision-reflective induction, clean behavior anchoring, and causal propagation, \textsc{TrojanWorld} establishes an end-to-end attack path from physical trigger perception, through corrupted imagination, to attacker-specified action selection. Evaluations on TD-MPC2, DreamerV3, and R2-Dreamer across four control benchmarks demonstrate that the attack can achieve targeted behavioral control with limited degradation of clean utility. More broadly, our results show that the security of world-model agents cannot be assessed solely at the policy or decision layer: the integrity of the internal predictive process itself is a critical part of the system's trust boundary. We hope these findings motivate more systematic end-to-end security evaluation and defense across world models.

\appendix

\bibliographystyle{plainurl}
\bibliography{main}

@article{ha2018worldmodels,
  title={World Models},
  author={Ha, David and Schmidhuber, J{\"u}rgen},
  journal={arXiv preprint arXiv:1803.10122},
  year={2018}
}

@article{lecun2022path,
  title={A Path Towards Autonomous Machine Intelligence},
  author={LeCun, Yann},
  journal={OpenReview},
  year={2022}
}

@article{ding2025understandingworld,
  title={Understanding World or Predicting Future? A Comprehensive Survey of World Models},
  author={Ding, Jingtao and Zhang, Yunke and Shang, Yu and Zhang, Yuheng and Zong, Zefang and Feng, Jie and Yuan, Yuan and Su, Hongyuan and Li, Nian and Sukiennik, Nicholas and Xu, Fengli and Li, Yong},
  journal={ACM Computing Surveys},
  year={2025}
}

@inproceedings{alonso2024diamond,
  title={Diffusion for World Modeling: Visual Details Matter in Atari},
  author={Alonso, Eloi and Jelley, Adam and Micheli, Vincent and Kanervisto, Anssi and Storkey, Amos and Pearce, Tim and Fleuret, Fran{\c{c}}ois},
  booktitle={Advances in Neural Information Processing Systems},
  volume={37},
  pages={58757--58791},
  year={2024}
}

@inproceedings{zhou2025dinowm,
  title={{DINO-WM}: World Models on Pre-trained Visual Features Enable Zero-shot Planning},
  author={Zhou, Gaoyue and Pan, Hengkai and LeCun, Yann and Pinto, Lerrel},
  booktitle={International Conference on Machine Learning},
  series={Proceedings of Machine Learning Research},
  volume={267},
  pages={79115--79135},
  year={2025}
}

@inproceedings{hansen2026newt,
  title={Learning Massively Multitask World Models for Continuous Control},
  author={Hansen, Nicklas and Su, Hao and Wang, Xiaolong},
  booktitle={International Conference on Learning Representations},
  year={2026}
}

@article{tan2026worldmodelsvla,
  title={Towards Generalist Embodied AI: A Survey on World Models for VLA Agents},
  author={Tan, Wentao and Zhu, Lei and Wang, Bowen and Xie, Enci and Ji, Baixu and Lin, Zengrong and Yang, Wenjie and Li, Jingjing and Shen, Heng Tao},
  journal={TechRxiv preprint},
  year={2026}
}

@inproceedings{hafner2020dreamer,
  title={Dream to Control: Learning Behaviors by Latent Imagination},
  author={Hafner, Danijar and Lillicrap, Timothy and Ba, Jimmy and Norouzi, Mohammad},
  booktitle={International Conference on Learning Representations},
  year={2020}
}

@article{hafner2025dreamerv3,
  title={Mastering Diverse Control Tasks Through World Models},
  author={Hafner, Danijar and Pasukonis, Jurgis and Ba, Jimmy and Lillicrap, Timothy},
  journal={Nature},
  volume={640},
  number={8059},
  pages={647--653},
  year={2025}
}

@inproceedings{hansen2022tdmpc,
  title={Temporal Difference Learning for Model Predictive Control},
  author={Hansen, Nicklas and Wang, Xiaolong and Su, Hao},
  booktitle={International Conference on Machine Learning},
  year={2022}
}

@inproceedings{hansen2024tdmpc2,
  title={TD-MPC2: Scalable, Robust World Models for Continuous Control},
  author={Hansen, Nicklas and Su, Hao and Wang, Xiaolong},
  booktitle={International Conference on Learning Representations},
  year={2024}
}

@inproceedings{morihira2026r2dreamer,
  title={R2-Dreamer: Redundancy-Reduced World Models Without Decoders or Augmentation},
  author={Morihira, Naoki and Nahar, Amal and Bharadwaj, Kartik and Kato, Yasuhiro and Hayashi, Akinobu and Harada, Tatsuya},
  booktitle={International Conference on Learning Representations},
  year={2026}
}

@inproceedings{tassa2018dmc,
  title={DeepMind Control Suite},
  author={Tassa, Yuval and Doron, Yotam and Muldal, Alistair and Erez, Tom and Li, Yazhe and de Las Casas, Diego and Budden, David and Abdolmaleki, Abbas and Merel, Josh and Lefrancq, Andrew and Lillicrap, Timothy and Riedmiller, Martin},
  booktitle={arXiv preprint arXiv:1801.00690},
  year={2018}
}

@inproceedings{yu2020metaworld,
  title={Meta-World: A Benchmark and Evaluation for Multi-Task and Meta Reinforcement Learning},
  author={Yu, Tianhe and Quillen, Deirdre and He, Zhanpeng and Julian, Ryan and Hausman, Karol and Finn, Chelsea and Levine, Sergey},
  booktitle={Conference on Robot Learning},
  year={2020}
}

@article{gu2017badnets,
  title={BadNets: Identifying Vulnerabilities in the Machine Learning Model Supply Chain},
  author={Gu, Tianyu and Dolan-Gavitt, Brendan and Garg, Siddharth},
  journal={arXiv preprint arXiv:1708.06733},
  year={2017}
}

@inproceedings{chou2023baddiffusion,
  title={How to Backdoor Diffusion Models?},
  author={Chou, Sheng-Yen and Chen, Pin-Yu and Ho, Tsung-Yi},
  booktitle={Proceedings of the IEEE/CVF Conference on Computer Vision and Pattern Recognition},
  pages={4015--4024},
  year={2023}
}

@inproceedings{liu2022badencoder,
  title={BadEncoder: Backdoor Attacks to Pre-trained Encoders in Self-Supervised Learning},
  author={Jia, Jinyuan and Liu, Yupei and Gong, Neil Zhenqiang},
  booktitle={IEEE Symposium on Security and Privacy},
  year={2022}
}

@inproceedings{kiourti2020trojdrl,
  title={TrojDRL: Trojan Attacks on Deep Reinforcement Learning Agents},
  author={Kiourti, Panagiota and Wardega, Kacper and Jha, Susmit and Li, Wenchao},
  booktitle={ACM/IEEE Design Automation Conference},
  year={2020}
}

@inproceedings{zhang2020rewardpoisoning,
  title={Adaptive Reward-Poisoning Attacks Against Reinforcement Learning},
  author={Zhang, Xuezhou and Ma, Yuzhe and Singla, Adish and Zhu, Xiaojin},
  booktitle={International Conference on Machine Learning},
  year={2020}
}

@inproceedings{rathbun2024sleepernets,
  title={SleeperNets: Universal Backdoor Poisoning Attacks Against Reinforcement Learning Agents},
  author={Rathbun, Ethan and Amato, Christopher and Oprea, Alina},
  booktitle={Advances in Neural Information Processing Systems},
  year={2024}
}

@inproceedings{zhan2026beat,
  title={BEAT: Visual Backdoor Attacks on VLM-Based Embodied Agents via Contrastive Trigger Learning},
  author={Zhan, Qiusi and Ha, Hyeonjeong and Yang, Rui and Xu, Sirui and Chen, Hanyang and Gui, Liang-Yan and Wang, Yu-Xiong and Zhang, Huan and Ji, Heng and Kang, Daniel},
  booktitle={International Conference on Learning Representations},
  year={2026}
}

@inproceedings{hafner2019planet,
  title={Learning Latent Dynamics for Planning from Pixels},
  author={Hafner, Danijar and Lillicrap, Timothy and Fischer, Ian and Villegas, Ruben and Ha, David and Lee, Honglak and Davidson, James},
  booktitle={International Conference on Machine Learning},
  year={2019}
}

@inproceedings{hu2025vpp,
  title={Video Prediction Policy: A Generalist Robot Policy with Predictive Visual Representations},
  author={Hu, Yucheng and Guo, Yanjiang and Wang, Pengchao and Chen, Xiaoyu and Wang, Yen-Jen and Zhang, Jianke and Sreenath, Koushil and Lu, Chaochao and Chen, Jianyu},
  booktitle={International Conference on Machine Learning},
  series={Proceedings of Machine Learning Research},
  volume={267},
  pages={24328--24346},
  year={2025}
}

@article{guo2026physcond,
  title={When World Models Dream Wrong: Physical-Conditioned Adversarial Attacks against World Models},
  author={Guo, Zhixiang and Liang, Siyuan and Balogh, Andras and Lunberry, Noah and Tu, Rong-Cheng and Jelasity, Mark and Tao, Dacheng},
  journal={arXiv preprint arXiv:2602.18739},
  year={2026}
}

@article{guo2026wmattack,
  title={{WMAttack}: Automated Attack Search for Adversarial Evaluation of World-Model Agents},
  author={Guo, Zhixiang and Liang, Siyuan and Fu, Shi and Guo, Cheng and Balogh, Andras and Jelasity, Mark and Tao, Dacheng},
  journal={arXiv preprint arXiv:2605.23220},
  year={2026}
}

@article{hu2026swaap,
  title={Stealthy World Model Manipulation via Data Poisoning},
  author={Hu, Yibin and Sun, Xiaolin and Zheng, Zizhan},
  journal={arXiv preprint arXiv:2606.18697},
  year={2026}
}

@article{duan2026trap,
  title={{TRAP}: Tail-aware Ranking Attack for World-Model Planning},
  author={Duan, Siyuan and Zhang, Ke and Luo, Xizhao},
  journal={arXiv preprint arXiv:2605.01950},
  year={2026},
}

@inproceedings{MyoSuite2022,
  title={MyoSuite: A Contact-rich Simulation Suite for Musculoskeletal Motor Control},
  author={Caggiano, Vittorio and Wang, Huawei and Durandau, Guillaume and Sartori, Massimo and Kumar, Vikash},
  booktitle={Proceedings of the 4th Annual Learning for Dynamics and Control Conference},
  series={Proceedings of Machine Learning Research},
  volume={168},
  pages={492--507},
  year={2022}
}

@misc{kannan2021robodesk,
  author={Kannan, Harini and Hafner, Danijar and Finn, Chelsea and Erhan, Dumitru},
  title={{RoboDesk}: A Multi-Task Reinforcement Learning Benchmark},
  year={2021},
  howpublished={Google Research software benchmark},
  url={https://github.com/google-research/robodesk}
}

@misc{brooks2024sora,
  title={Video Generation Models as World Simulators},
  author={Brooks, Tim and Peebles, Bill and Holmes, Connor and DePue, Will and Guo, Yufei and Jing, Li and Schnurr, David and Taylor, Joe and Luhman, Troy and Luhman, Eric and Ng, Clarence and Wang, Ricky and Ramesh, Aditya},
  year={2024},
  howpublished={OpenAI Technical Report},
  url={https://openai.com/index/video-generation-models-as-world-simulators/}
}

@misc{ye2026dreamzero,
  title={World Action Models are Zero-shot Policies},
  author={Ye, Seonghyeon and Ge, Yunhao and Zheng, Kaiyuan and Gao, Shenyuan and Yu, Sihyun and Kurian, George and Indupuru, Suneel and Tan, You Liang and Zhu, Chuning and Xiang, Jiannan and Malik, Ayaan and Lee, Kyungmin and Liang, William and Ranawaka, Nadun and Gu, Jiasheng and Xu, Yinzhen and Wang, Guanzhi and Hu, Fengyuan and Narayan, Avnish and Bjorck, Johan and Wang, Jing and Kim, Gwanghyun and Niu, Dantong and Zheng, Ruijie and Xie, Yuqi and Wu, Jimmy and Wang, Qi and Julian, Ryan and Xu, Danfei and Du, Yilun and Chebotar, Yevgen and Reed, Scott and Kautz, Jan and Zhu, Yuke and Fan, Linxi and Jang, Joel},
  year={2026},
  note={arXiv preprint arXiv:2602.15922}
}

@inproceedings{wang2019neuralcleanse,
  title={Neural Cleanse: Identifying and Mitigating Backdoor Attacks in Neural Networks},
  author={Wang, Bolun and Yao, Yuanshun and Shan, Shawn and Li, Huiying and Viswanath, Bimal and Zheng, Haitao and Zhao, Ben Y.},
  booktitle={IEEE Symposium on Security and Privacy},
  pages={707--723},
  year={2019}
}

@inproceedings{liu2018finepruning,
  title={Fine-Pruning: Defending Against Backdooring Attacks on Deep Neural Networks},
  author={Liu, Kang and Dolan-Gavitt, Brendan and Garg, Siddharth},
  booktitle={Research in Attacks, Intrusions, and Defenses},
  pages={273--294},
  year={2018}
}

@article{deboer2005cem,
  title={A Tutorial on the Cross-Entropy Method},
  author={de Boer, Pieter-Tjerk and Kroese, Dirk P. and Mannor, Shie and Rubinstein, Reuven Y.},
  journal={Annals of Operations Research},
  volume={134},
  number={1},
  pages={19--67},
  year={2005},
}

@article{tsochantaridis2005largemargin,
  title={Large Margin Methods for Structured and Interdependent Output Variables},
  author={Tsochantaridis, Ioannis and Joachims, Thorsten and Hofmann, Thomas and Altun, Yasemin},
  journal={Journal of Machine Learning Research},
  volume={6},
  number={50},
  pages={1453--1484},
  year={2005}
}

@inproceedings{shrivastava2016ohem,
  title={Training Region-Based Object Detectors with Online Hard Example Mining},
  author={Shrivastava, Abhinav and Gupta, Abhinav and Girshick, Ross},
  booktitle={Proceedings of the IEEE Conference on Computer Vision and Pattern Recognition},
  pages={761--769},
  year={2016}
}

@inproceedings{czarnecki2019policydistillation,
  title={Distilling Policy Distillation},
  author={Czarnecki, Wojciech M. and Pascanu, Razvan and Osindero, Simon and Jayakumar, Siddhant and Swirszcz, Grzegorz and Jaderberg, Max},
  booktitle={Proceedings of the Twenty-Second International Conference on Artificial Intelligence and Statistics},
  series={Proceedings of Machine Learning Research},
  volume={89},
  pages={1331--1340},
  year={2019}
}

@inproceedings{wu2023daydreamer,
  title={{DayDreamer}: World Models for Physical Robot Learning},
  author={Wu, Philipp and Escontrela, Alejandro and Hafner, Danijar and Abbeel, Pieter and Goldberg, Ken},
  booktitle={Conference on Robot Learning},
  series={Proceedings of Machine Learning Research},
  volume={205},
  pages={2226--2240},
  year={2023}
}

@inproceedings{liu2018trojaning,
  title={Trojaning Attack on Neural Networks},
  author={Liu, Yingqi and Ma, Shiqing and Aafer, Yousra and Lee, Wen-Chuan and Zhai, Juan and Wang, Weihang and Zhang, Xiangyu},
  booktitle={Network and Distributed System Security Symposium},
  year={2018}
}

@inproceedings{bagdasaryan2021blind,
  title={Blind Backdoors in Deep Learning Models},
  author={Bagdasaryan, Eugene and Shmatikov, Vitaly},
  booktitle={30th USENIX Security Symposium (USENIX Security 21)},
  pages={1505--1521},
  year={2021}
}

@article{li2024backdoorsurvey,
  title={Backdoor Learning: A Survey},
  author={Li, Yiming and Jiang, Yong and Li, Zhifeng and Xia, Shu-Tao},
  journal={IEEE Transactions on Neural Networks and Learning Systems},
  volume={35},
  number={1},
  pages={5--22},
  year={2024},
}

@inproceedings{yuan2024shine,
  title={{SHINE}: Shielding Backdoors in Deep Reinforcement Learning},
  author={Yuan, Zhuowen and Guo, Wenbo and Jia, Jinyuan and Li, Bo and Song, Dawn},
  booktitle={Proceedings of the 41st International Conference on Machine Learning},
  pages={57887--57904},
  volume={235},
  series={Proceedings of Machine Learning Research},
  year={2024}
}

@article{tang2025deep,
  title={Deep reinforcement learning for robotics: A survey of real-world successes},
  author={Tang, Chen and Abbatematteo, Ben and Hu, Jiaheng and Chandra, Rohan and Mart{\'\i}n-Mart{\'\i}n, Roberto and Stone, Peter},
  journal={Annual Review of Control, Robotics, and Autonomous Systems},
  volume={8},
  number={1},
  pages={153--188},
  year={2025},
  publisher={Annual Reviews}
}

@inproceedings{wang2025modelsupply,
  title={Model supply chain poisoning: Backdooring pre-trained models via embedding indistinguishability},
  author={Wang, Hao and Guo, Shangwei and He, Jialing and Liu, Hangcheng and Zhang, Tianwei and Xiang, Tao},
  booktitle={Proceedings of the ACM on Web Conference 2025},
  pages={840--851},
  year={2025}
}

@article{huang2024suprte,
  title={SupRTE: Suppressing backdoor injection in federated learning via robust trust evaluation},
  author={Huang, Wenkai and Li, Gaolei and Yi, Xiaoyu and Li, Jianhua and Zhao, Chengcheng and Yin, Ying},
  journal={IEEE Intelligent Systems},
  volume={39},
  number={5},
  pages={66--77},
  year={2024},
  publisher={IEEE}
}

@article{huang2025silent,
  title={Silent penetrator: Breaching cross-domain federated fine-tuning via feature shift-induced backdoor},
  author={Huang, Wenkai and Li, Gaolei and Chen, Mingzhe and Li, Jianhua and Zhu, Haojin},
  journal={IEEE Transactions on Information Forensics and Security},
  volume={20},
  pages={7106--7120},
  year={2025},
  publisher={IEEE}
}

@inproceedings{bruce2024genie,
  title={Genie: Generative interactive environments},
  author={Bruce, Jake and Dennis, Michael D and Edwards, Ashley and Parker-Holder, Jack and Shi, Yuge and Hughes, Edward and Lai, Matthew and Mavalankar, Aditi and Steigerwald, Richie and Apps, Chris and others},
  booktitle={Forty-first international conference on machine learning},
  year={2024}
}

@inproceedings{yang2024unisim,
  title={Learning Interactive Real-World Simulators},
  author={Yang, Sherry and Du, Yilun and Ghasemipour, Seyed Kamyar Seyed and Tompson, Jonathan and Kaelbling, Leslie Pack and Schuurmans, Dale and Abbeel, Pieter},
  booktitle={International Conference on Learning Representations},
  year={2024}
}

@inproceedings{cheng2024lotus,
  title={Lotus: Evasive and resilient backdoor attacks through sub-partitioning},
  author={Cheng, Siyuan and Tao, Guanhong and Liu, Yingqi and Shen, Guangyu and An, Shengwei and Feng, Shiwei and Xu, Xiangzhe and Zhang, Kaiyuan and Ma, Shiqing and Zhang, Xiangyu},
  booktitle={2024 IEEE/CVF Conference on Computer Vision and Pattern Recognition (CVPR)},
  pages={24798--24809},
  year={2024},
  organization={IEEE}
}

@inproceedings{wang2024badagent,
  title={Badagent: Inserting and activating backdoor attacks in llm agents},
  author={Wang, Yifei and Xue, Dizhan and Zhang, Shengjie and Qian, Shengsheng},
  booktitle={Proceedings of the 62nd Annual Meeting of the Association for Computational Linguistics (Volume 1: Long Papers)},
  pages={9811--9827},
  year={2024}
}

@article{liu2025pre,
  title={Pre-trained trojan attacks for visual recognition},
  author={Liu, Aishan and Liu, Xianglong and Zhang, Xinwei and Xiao, Yisong and Zhou, Yuguang and Liang, Siyuan and Wang, Jiakai and Cao, Xiaochun and Tao, Dacheng},
  journal={International Journal of Computer Vision},
  volume={133},
  number={6},
  pages={3568--3585},
  year={2025},
  publisher={Springer}
}

@inproceedings{liang2024badclip,
  title={Badclip: Dual-embedding guided backdoor attack on multimodal contrastive learning},
  author={Liang, Siyuan and Zhu, Mingli and Liu, Aishan and Wu, Baoyuan and Cao, Xiaochun and Chang, Ee-Chien},
  booktitle={Proceedings of the IEEE/CVF conference on computer vision and pattern recognition},
  pages={24645--24654},
  year={2024}
}

@inproceedings{liang2025revisiting,
  title={Revisiting Backdoor Attacks against Large Vision-Language Models from Domain Shift},
  author={Liang, Siyuan and Liang, Jiawei and Pang, Tianyu and Du, Chao and Liu, Aishan and Zhu, Mingli and Cao, Xiaochun and Tao, Dacheng},
  booktitle={Proceedings of the Computer Vision and Pattern Recognition Conference},
  pages={9477--9486},
  year={2025}
}

@article{liu2024compromising,
  title={Compromising embodied agents with contextual backdoor attacks},
  author={Liu, Aishan and Zhou, Yuguang and Liu, Xianglong and Zhang, Tianyuan and Liang, Siyuan and Wang, Jiakai and Pu, Yanjun and Li, Tianlin and Zhang, Junqi and Zhou, Wenbo and others},
  journal={arXiv preprint arXiv:2408.02882},
  year={2024}
}

@inproceedings{zhang2024towards,
  title={Towards robust physical-world backdoor attacks on lane detection},
  author={Zhang, Xinwei and Liu, Aishan and Zhang, Tianyuan and Liang, Siyuan and Liu, Xianglong},
  booktitle={Proceedings of the 32nd ACM International Conference on Multimedia},
  pages={5131--5140},
  year={2024}
}

@article{liu2025natural,
  title={Natural Reflection Backdoor Attack on Vision Language Model for Autonomous Driving},
  author={Liu, Ming and Liang, Siyuan and Howlader, Koushik and Wang, Liwen and Tao, Dacheng and Zhang, Wensheng},
  journal={arXiv preprint arXiv:2505.06413},
  year={2025}
}

@article{zhu2024breaking,
  title={Breaking the false sense of security in backdoor defense through re-activation attack},
  author={Zhu, Mingli and Liang, Siyuan and Wu, Baoyuan},
  journal={Advances in Neural Information Processing Systems},
  volume={37},
  pages={114928--114964},
  year={2024}
}

@article{xun2025robust,
  title={Robust Anti-Backdoor Instruction Tuning in LVLMs},
  author={Xun, Yuan and Liang, Siyuan and Jia, Xiaojun and Liu, Xinwei and Cao, Xiaochun},
  journal={arXiv preprint arXiv:2506.05401},
  year={2025}
}

\end{document}